\documentclass[twoside]{article}

\usepackage[accepted]{aistats2023}
\usepackage[round]{natbib}

\usepackage[utf8]{inputenc} 
\usepackage[T1]{fontenc}    
\usepackage{hyperref}       
\usepackage{url}            
\usepackage{booktabs}       
\usepackage{amsfonts}       
\usepackage{nicefrac}       
\usepackage{microtype}      
\usepackage{xcolor}         
\usepackage{natbib}
\usepackage{soul}

\usepackage{wrapfig}
\usepackage{adjustbox}
\usepackage{amsmath}
\usepackage{amssymb}
\usepackage{epsfig}
\usepackage{graphics,amsopn,amstext,color,enumerate,bm}
\usepackage{float,epstopdf}
\usepackage{mathtools}
\usepackage{svg}
\usepackage{multirow}
\usepackage{ulem}
\usepackage{graphicx}
\usepackage{subfigure}

\newcommand{\bA}{{\mathbf{A}}}

\newcommand{\by}{{\mathbf{y}}}

\newcommand{\bx}{{\mathbf{x}}}
\newcommand{\bz}{{\mathbf{z}}}
\newcommand{\ba} {{\mathbf{a}}}
\newcommand{\bb}{{\mathbf{b}}}
\newcommand{\bc}{{\mathbf{c}}}
\newcommand{\bd}{{\mathbf{d}}}
\newcommand{\be}{{\mathbf{e}}}

\newcommand{\bv}{{\mathbf{v}}}
\newcommand{\bu}{{\mathbf{u}}}
\newcommand{\bP}{{\mathbf{P}}}
\newcommand{\bq}{{\mathbf{q}}}
\newcommand{\bW}{{\mathbf{W}}}

\newcommand{\bS}{{\mathbf{S}}}
\newcommand{\bD}{{\mathbf{D}}}

\newcommand{\btheta}{{\bm{\theta}}}

\renewcommand{\st}{{\rm s.t.}}

\usepackage{cleveref}
\crefname{algorithm}{Algorithm}{Algorithms}
\crefname{assumption}{Assumption}{Assumptions}
\crefname{equation}{Eq.}{Eqs.}
\crefname{figure}{Fig.}{Figs.}
\crefname{table}{Table}{Tables}
\crefname{section}{Sec.}{Secs.}
\crefname{theorem}{Theorem}{Theorems}
\crefname{lemma}{Lemma}{Lemmas}
\crefname{proposition}{Proposition}{Propositions}
\crefname{definition}{Definition}{Definitions}
\crefname{corollary}{Corollary}{Corollaries}
\crefname{rem}{Remark}{Remarks}
\crefname{example}{Example}{Examples}
\crefname{appendix}{Appendix}{Appendices}

\DeclareMathOperator*{\argmin}{argmin}
\DeclareMathOperator*{\argmax}{argmax}

\usepackage{algpseudocode, algorithm}
\usepackage[noend]{algcompatible}

\newtheorem{theorem}{Theorem}
\newtheorem{lemma}[theorem]{Lemma} 
 
\newtheorem{remark}[theorem]{Remark}

\newtheorem{definition}[theorem]{Definition}

\begin{document}

%

%

\twocolumn[

\aistatstitle{Improving Adversarial Robustness via Joint Classification and Multiple Explicit Detection Classes}

\aistatsauthor{ Sina Baharlouei \And Fatemeh Sheikholeslami\footnotemark[1] \And  Meisam Razaviyayn \And Zico Kolter}

\aistatsaddress{USC\\baharlou@usc.edu\\ \And Amazon Alexa AI \\shfateme@amazon.com\And USC\\razaviya@usc.edu \And Bosch Center for AI, CMU\\zkolter@cs.cmu.edu}
]

\begin{abstract}
\vspace{-0.4cm}
This work concerns the development of deep networks that are certifiably robust to adversarial attacks. Joint robust classification-detection was recently introduced as a  certified defense mechanism, where adversarial examples are either correctly classified \emph{or} assigned to the ``abstain'' class. In this work, we show that such a provable framework can benefit by extension to networks with \emph{multiple} explicit abstain classes, where the adversarial examples are adaptively assigned to those. We show that na\"ively adding multiple abstain classes can lead to ``model degeneracy'', then we propose a regularization approach and a training method to counter this degeneracy by promoting full use of the multiple abstain classes.  Our experiments demonstrate that the proposed approach consistently achieves favorable standard vs. robust verified accuracy tradeoffs, outperforming state-of-the-art algorithms for various choices of number of abstain classes. Our code is available at \url{https://github.com/sinaBaharlouei/MultipleAbstainDetection}.\footnotetext[1]{The work of FS was done when FS was with the Bosch Center for AI.}
\end{abstract}
\vspace{-0.2cm}
\section{Introduction}
\vspace{-0.4cm}
Deep Neural Networks (DNNs) have revolutionized many machine learning tasks such as image processing~\citep{krizhevsky2012imagenet, zhu2021webface260m} and speech recognition~\citep{graves2013speech, nassif2019speech}. 
However, despite their superior performance, DNNs are highly vulnerable to adversarial attacks and perform poorly on out-of-distributions samples~\citep{goodfellow2014explaining, liang2017enhancing, yuan2019adversarial}. 
To address the vulnerability of DNNs to adversarial attacks, the community has designed various defense mechanisms against such attacks~\citep{papernot2016distillation,jang2019adversarial, goldblum2020adversarially, madry2017towards,huang2021dair}. These mechanisms provide robustness against certain types of attacks, such as the Fast Gradient Sign Method (FGSM)~\citep{szegedy2013intriguing, goodfellow2014explaining}. However, the overwhelming majority of these defense mechanisms are highly ineffective against more complex attacks such as adaptive and brute-force methods~\citep{tramer2020adaptive, carlini2017towards}. This ineffectiveness necessitates: 1) the design of rigorous verification approaches that can measure the robustness of a given network; 2) the development of  defense mechanisms that are \emph{verifiably} robust against \textit{any} attack strategy within the class of permissible attack strategies. 

To verify the robustness of a given network against \textit{any} attack in a reasonable set of permissible attacks (e.g. $\ell_p$-norm ball around the given input data), one needs to solve a hard non-convex optimization problem (see, e.g., Problem~\eqref{eq: verification_original_problem} in this paper). Consequently, exact verifiers, such as the ones developed in~\citep{tjeng2017evaluating, xiao2018training}, are not scalable to large networks. To develop scalable verifiers, the community turn to ``inexact" verifiers which can only verify a subset of perturbations to the input data that the network can defend against successfully. These verifiers  typically rely on  tractable lower bounds for the verification optimization problem. \citet{gowal2018effectiveness} finds such a  lower-bound by \textit{interval bound propagation (IBP)}, which is essentially an efficient convex relaxation of the constraint sets in the verification problem. 
Despite its simplicity, this approach demonstrates relatively superior performance compared to  prior works. 

IBP-CROWN~\citep{zhang2019towards} combines IBP with novel linear relaxations to have a tighter approximation than standalone IBP. $\beta$-Crown~\citep{wang2021beta} utilizes a branch-and-bound technique combined with the linear bounds in IBP-CROWN to tighten the relaxation gap further. While $\beta$-Crown demonstrates a tremendous performance gain over other verifiers such as~\citet{zhang2019towards, fazlyab2019efficient, lu2019neural}, it cannot be used as a tool in large-scale training procedures due to its computationally expensive branch-and-bound search. One can adopt a composition of certified architectures to enhance the performance of the obtained model on both natural and adversarial accuracy~\citep{muller2021certify, horvath2022robust}.

Another line of work for enhancing the performance of certifiably robust neural networks relies on the idea of learning a detector alongside the classifier to capture adversarial samples. Instead of trying to classify adversarial images correctly, these works design a \textit{detector} to determine whether a given sample is natural/in-distribution  or a crafted attack/out-of-distribution. 
\citet{chen2020robust} train the detector on both in-distribution and out-of-distribution samples to learn a detector distinguishing these samples. \citet{hendrycks2016baseline} develops a method based on a simple observation that, for real samples, the output of softmax layer is  closer to $0$ or $1$ compared to out-of-distribution and adversarial examples where the softmax output entries are distributed more uniformly. 
\citet{devries2018learning, sheikholeslami2020minimum, stutz2020confidence} learn uncertainty regions around actual samples where the network prediction remains the same. 
Interestingly, this approach does not require out-of-distribution samples during training. 
Other approaches such as deep generative models~\citep{ren2019likelihood}, self-supervised and ensemble methods~\citep{vyas2018out, chen2021understanding} are also used to learn out-of-distribution samples. However, typically these methods are vulnerable to adversarial attacks and can be easily fooled by carefully designed out-of-distribution images~\citep{fort2022adversarial} as discussed in~\citet{tramer2022detecting}. 
A more resilient approach is to jointly learn the detector and the classifier~\citep{laidlaw2019playing, sheikholeslamiprovably, chen2021revisiting} by adding an auxiliary \textit{abstain} output class capturing adversarial samples. 

Building on these prior works, this paper develops a framework for detecting adversarial examples using multiple abstain classes. We observe that  na\"ively adding multiple abstain classes (in the existing framework of \citet{sheikholeslamiprovably}) results in a model degeneracy phenomenon where all adversarial examples are assigned to a small fraction of abstain classes (while other abstain classes are not utilized). To resolve this issue, we propose a novel regularizer and a training procedure to balance the assignment of adversarial examples to abstain classes. 
Our experiments demonstrate that utilizing multiple abstain classes in conjunction with the proper regularization enhances the robust verified accuracy  on adversarial examples while maintaining the standard accuracy of the classifier. 

\noindent\textbf{Challenges and Contribution.} We propose a framework for training and verifying robust neural nets with multiple detection classes. The resulting optimization problems for training and verifying such networks is a \textit{constrained min-max} optimization problem over a probability simplex that is more challenging from an optimization perspective than the problems associated with networks with no or single detection classes. We devise an efficient  algorithm for this problem.
Furthermore, having multiple detectors leads to the ``model degeneracy" phenomenon, where not all detection classes are utilized. To prevent model degeneracy and \text{to avoid tuning the number of network detectors}, we introduce a regularization mechanism guaranteeing that all detectors contribute to detecting adversarial examples to the extent possible. We propose convergent algorithms for the verification (and training)  problems using proximal gradient descent with Bregman divergence. Compared to networks with a single detection class, our experiments show that we enhance the robust verified accuracy by more than $5\%$ and $2\%$ on CIFAR-10 and MNIST datasets, respectively, for various perturbation sizes.

\noindent\textbf{Roadmap.} In section~\ref{sec:background} we review interval bound propagation (IBP) and $\beta$-crown as two existing efficient  methods for verifying the performance of multi-layer neural networks against adversarial attacks. 
We  discuss how to train and verify joint classifier and detector networks (with a single abstain class)  based on these two approaches. Section~\ref{sec:verification} is dedicated to the motivation and procedure of joint verification and classification of neural networks with \textbf{multiple} abstain classes. In particular, we extend IBP and $\beta$-crown verification procedures to networks with multiple detection classes. In section~\ref{sec:training}, we show how to train neural networks with multiple detection classes via IBP procedure. However, we show that the performance of the trained network cannot be  improved by only increasing the number of detection classes due to  ``model degeneracy" (a phenomenon that happens when multiple detectors behave very similarly and identify the same adversarial examples). To avoid model degeneracy and to automatically/implicitly tune the number of detection classes, we introduce a regularization mechanism such that all detection classes are used in balance.
\vspace{-3mm}
\section{Background}
\label{sec:background}
\vspace{-0.2cm}
\subsection{Verification of feedforward neural networks}
\vspace{-0.2cm}
Consider an $L$-layer feedforward neural network with $\{\bW_i, \bb_i\}$ denoting the weight and bias parameters associated with layer~$i$, 
and let~$\sigma_i(\cdot)$ denote the activation function applied at layer~$i$. Throughout the paper, we assume the activation function is the same for all hidden layers, i.e., $\sigma_i(\cdot)= \sigma(\cdot) = \textrm{ReLU}(\cdot),\;\forall i = 1,\ldots, L-1$. Thus, our neural network can  be described as
\begin{equation*} 
\small
    \bz_i = \sigma(\bW_i \bz_{i-1} + \bb_i) \: \: \: \forall i \in [L-1], \: \: \bz_L = \bW_{L} \bz_{L-1} + \bb_L,
\end{equation*}
where $\bz_0 = \bx$ is the input to the neural network and $\bz_i$ is the output of layer $i$ and $[N]$ denotes the set $\{1, \dots, N\}$. Note that the activation function is not applied at the last layer. Further, we use $[\bz]_i$ to denote the $i$-th element of the vector $\bz$. 
We consider a supervised classification task where~$\bz_L$ represents the logits. To explicitly show the dependence of $\bz_L$ on the input data, we use the notation $\bz_L(\bx)$ to denote logit values when~$\bx$ is used as the input data point. 

Given an input $\bx_0$ with the ground-truth label $y$, and a perturbation set~$\mathcal{C}(\bx_0, \epsilon)$ (e.g. $ \mathcal{C}(\bx_0, \epsilon)= \{\bx \, | \, \| \bx - \bx_0 \|_{\infty} \leq \epsilon \}$), the network is provably robust against adversarial attacks on~$\bx_0$ if 
\vspace{-4mm}
\begin{equation}
\label{eq: verification_original_problem}
0 \leq \min_{\bx \in \mathcal{C}(\bx_0, \epsilon)} \bc_{yk} ^T \bz_{L} (\bx), \quad \forall \: k \neq y,
\end{equation}
where $\bc_{yk} = \be_y - \be_k$ with $\be_k$ (resp. $\be_y$) denoting the standard unit vector whose $k$-th row (resp. $y$-th row) is~$1$ and the other entries are zero. 
Condition~\eqref{eq: verification_original_problem} implies that the logit score of the network for the true label $y$ is always greater than that of any other label~$k$ for all $\bx \in \mathcal{C}(\bx_0, \epsilon)$. Thus, the network will correctly classify all the points inside $\mathcal{C}(\bx_0, \epsilon)$. 
The objective function in~\cref{eq: verification_original_problem} is non-convex when~$L\geq 2$. It is customary in many works to move the non-convexity of the problem to the constraint set and reformulate~\cref{eq: verification_original_problem} as
\begin{equation}
 \label{eq: verification_problem_z}
0 \leq \min_{\bz \in \mathcal{Z}(\bx_0, \epsilon)} \bc_{yk} ^T \bz, \quad \quad \forall \: k \neq y,
\end{equation}
where $\mathcal{Z}(\bx_0, \epsilon) = \{ \bz \, | \, \bz = \bz_L(\bx) \textrm{ for some } \bx \in \mathcal{C}(\bx_0, \epsilon)\}$. This verification problem has a linear objective function and a non-convex constraint set. Since both problems~\eqref{eq: verification_original_problem} and \eqref{eq: verification_problem_z} are non-convex, existing works proposed efficiently computable lower-bounds for the optimal objective value of them. For example, \citet{gowal2018effectiveness, wong2018provable} utilize convex relaxation, while \citet{tjeng2017evaluating, wang2021beta} rely on mixed integer programming and branch-and-bound to find lower-bounds for the optimal objective value of~\eqref{eq: verification_problem_z}. In what follows, we explain two popular and relatively successful approaches for solving the verification problem~\eqref{eq: verification_original_problem} (or equivalently~\eqref{eq: verification_problem_z}) in detail.
\vspace{-2mm}
\subsection{Verification of neural networks via IBP} \label{subsec:IBP}
\vspace{-2mm}
Interval Bound Propagation (IBP) of~\citet{gowal2018effectiveness} tackles problem~\eqref{eq: verification_problem_z} by convexification of the constraint set $\mathcal{Z}(\bx_0, \epsilon)$ to its convex hypercube super-set~$[\underline{\bz}(\bx_0), \bar{\bz}(\bx_0)]$, i.e., $\mathcal{Z}(\bx_0, \epsilon) \subseteq [\underline{\bz}(\bx_0), \bar{\bz}(\bx_0)]$.  After this relaxation, problem~\eqref{eq: verification_problem_z} can be lower-bounded by the convex problem:
%
\begin{equation}
\label{eq: verification_IBP_problem}
\min_{\underline{\bz}(\bx_0) \leq \bz \leq \bar{\bz}(\bx_0)} \bc_{yk} ^T \bz
\end{equation}
The  upper- and lower- bounds $\underline{\bz}(\bx_0)$ and $\bar{\bz}(\bx_0)$  are obtained by recursively finding the convex relaxation of the image of the set $\mathcal{C}(\bx_0,\epsilon)$ at each layer of the network. In particular, for the adversarial set~$\mathcal{C}(\bx_0, \epsilon)= \{\bx \, | \, \| \bx - \bx_0 \|_{\infty} \leq \epsilon \}$,  we start from $\underline{\bz}_0(\bx_0) = \bx_0 - \epsilon \mathbf{1}$ and $\bar{\bz}_0(\bx_0) = \bx_0 + \epsilon \mathbf{1}$. Then, the lower-bound $\underline{\bz}_L(\bx_0)$ and upper-bound $\bar{\bz}_L(\bx_0)$ are computed by the recursions for all $i \in [L]$: 
\begin{equation}\label{eq:IBP}
\begin{split}
\small
    \bar{\bz}_i(\bx_0) &= \sigma(\bW_i^T \frac{\bar{\bz}_{i-1} + \underline{\bz}_{i-1}}{2} + |\bW_i^T|\frac{\bar{\bz}_{i-1} - \underline{\bz}_{i-1}}{2}), \\
    \underline{\bz}_i(\bx_0) &= \sigma(\bW_i^T \frac{\bar{\bz}_{i-1} + \underline{\bz}_{i-1}}{2} - |\bW_i^T|\frac{\bar{\bz}_{i-1} - \underline{\bz}_{i-1}}{2}). 
\end{split}
\end{equation}
Note that $|\bW|$ denotes the element-wise absolute value of matrix $\bW$. One of the main advantages of IBP is its efficient computation: verification of a given input only requires  two forward passes for finding the lower and upper bounds, followed by a  linear programming.
\vspace{-2mm}
\subsection{Verification of neural networks via \texorpdfstring{$\beta$}--Crown}
\vspace{-2mm}
Despite its simplicity, IBP-based verification comes with a certain limitation, namely the looseness of its layer-by-layer bounds of the input. To overcome this limitation, tighter verification methods have been proposed in the literature~\citep{singh2018fast, zhang2019towards, dathathri2020enabling,wang2021beta}. 
Among these, $\beta$-crown~\citep{wang2021beta} utilizes the branch-and-bound technique to generalize and improve the  IBP-CROWN proposed in~\citet{zhang2019towards}. 
Let $\underline{\bz}_i$ and $\bar{\bz}_i$ be the estimated element-wise lower-bound and upper-bounds for the pre-activation value of $\bz_i$, i.e., $
    \underline{\bz}_i \leq \bz_i \leq \bar{\bz}_i$, where  these lower and upper bounds are obtained by the method in~\citet{zhang2019towards}.
Let $\hat{\bz}_i$ be the value we obtain by applying ReLU function to $\bz_i$.
A neuron is called unstable if its sign after applying 
ReLU activation cannot be determined based on only knowing the corresponding lower and upper bounds. That is, a neuron is unstable if $\underline{\bz}_i < 0 < \bar{\bz}_i$. For \textbf{stable} neurons, no relaxation is needed to enforce convexity of $\sigma(\bz)$ (since the neuron operates in a linear regime). On the other hand, given an unstable neuron, they use a branch-and-bound (BAB) approach to split the input range of the neuron into two sub-domains $\mathcal{C}_{il} = \{\bx \in \mathcal{C}(\bx_0, \epsilon) | \: \hat{z}_i \leq 0 \}$ and $\mathcal{C}_{iu} = \{\bx \in \mathcal{C}(\bx_0, \epsilon) | \: \hat{z}_i > 0 \}$. The neuron operates linearly within each subdomain. Thus we can verify each subdomain separately. If we have $N$ unstable nodes, 
the BAB algorithm requires the investigation of $2^N$ sub-domains in the worst case. $\beta$-Crown  proposes a heuristic for traversing all these subdomains: The higher the absolute value of the corresponding lower bound of a node is, the sooner the verifier visits it. For verifying each sub-problem, \citet{wang2021beta} proposed a  lower-bounded which requires solving a maximization problem over two parameters $\boldsymbol{\alpha}$ and $\boldsymbol{\beta}$:
\begin{align}
\label{eq: beta_crown_subproblem}
    & \min_{\bz \in \mathcal{Z}(\bx_0, \epsilon)} \bc_{yk}^T \bz  \geq \max_{\boldsymbol{\alpha}, \boldsymbol{\beta}} g(\bx, \boldsymbol{\alpha}, \boldsymbol{\beta}) \nonumber \\ & \textrm{where} \quad g(\bx, \boldsymbol{\alpha}, \boldsymbol{\beta}) = (\ba + \bP_{\boldsymbol{\alpha}} \boldsymbol{\beta})^T \bx + \bq_{\boldsymbol{\alpha}}^T \boldsymbol{\beta} + \bd_{\boldsymbol{\alpha}}. 
\end{align}
Here, the matrix $\bP$ and the vectors $\bq, \ba$ and $\bd$ are functions of $\bW_i, \bb_i, \underline{\bz}_i, \bar{\bz}_i, \boldsymbol{\alpha},$ and $\boldsymbol{\beta}$ parameters. 
See Appendix~\ref{appendix: beta_crown} 
for the precise definition of $g$.
Notice that any choice of $(\boldsymbol{\alpha}, \boldsymbol{\beta})$ provides a valid lower bound for verification. However, optimizing $\boldsymbol{\alpha}$ and $\boldsymbol{\beta}$ in \eqref{eq: beta_crown_subproblem} leads to a tighter bound. 

\vspace{-2mm}
\subsection{Training a joint robust classifier and detector}
\vspace{-2mm}
\citet{sheikholeslamiprovably} improves the performance tradeoff  on natural and adversarial examples by introducing an auxiliary class for detecting adversarial examples. If this auxiliary class is selected as the  output, the network ``abstains" from declaring any of the original $K$ classes for the given input. Let $a$ be the abstain class. The classification network performs correctly on an  adversarial image  if it is classified correctly as the class of the original (unperturbed) image (similar to robust networks without detectors) or it is classified as the abstain class (detected as an adversarial example). Hence, for input image $(\mathbb{x}_0,y)$ the network is verified against a certain class $k\neq y$  if
\begin{equation}
 \label{eq: verification_original_problem1}
0 \leq  \min_{\bz \in \mathcal{Z}(\bx_0, \epsilon)} \max (\bc_{yk} ^T \bz, \bc_{ak}^T \bz),
\end{equation}
i.e., if  the score of the true label $y$ or the score of the abstain class $a$ is larger than the score of class~$k$. 

To train a neural network that can jointly detect and classify a dataset of images, \citet{sheikholeslamiprovably} relies on the loss function of the form:
\begin{align}
\small
L_{\textrm{Total}} = L_{\textrm{Robust}} + \lambda_1 L_{\textrm{Robust}}^{\textrm{Abstain}}  +  \lambda_2 L_{\textrm{Natural}},
\label{eq: loss_abstain}
\end{align}
where the term $L_{\textrm{Natural}}$ denotes the natural loss  when no adversarial examples are considered. More precisely, $L_{\textrm{Natural}} = \frac{1}{n} \sum_{i=1}^n \ell_{\textrm{xent}} \big(\bz_L(\bx_i), y_i\big)$,
%
where $\ell_{\textrm{xent}}$ is the standard cross-entropy loss. 
 The term $L_{\textrm{Robust}}$ in \eqref{eq: loss_abstain} represents the worst-case adversarial loss used in \citep{madry2017towards}, without considering the abstain class. Precisely, 
\vspace{-2mm}
\begin{align*} 
\small
L_{\textrm{Robust}} =  & \max_{\boldsymbol{\delta}_1, \dots, \boldsymbol{\delta}_n} \: \: \frac{1}{n} \sum_{i=1}^n \ell_{\textrm{xent}} \big(\bz_L(\bx_i + \boldsymbol{\delta}_i),  y_i\big) \\ &\st \: \: \|\boldsymbol{\delta}_i \|_{\infty} \leq \epsilon, \;\;\forall i=1,\ldots, n.
\end{align*}
Finally, the Robust-Abstain loss $L_{\textrm{Robust}}^{\textrm{Abstain}}$  is the minimum of the detector and the classifier losses:
\small
\begin{equation}\label{eq: robust_abstain_loss}
\begin{split}
\small
    L_{\textrm{Robust}}^{\textrm{Abstain}} = &\max_{\boldsymbol{\delta}_1, \dots, \boldsymbol{\delta}_n} \quad \frac{1}{n} \sum_{i=1}^n \min \Big(  \ell_{\textrm{xent}} \big(\bz_L(\bx_i + \boldsymbol{\delta}_i),  y_i \big),  \\ 
    & \hspace{3cm} \ell_{\textrm{xent}} \big(\bz_L(\bx_i + \boldsymbol{\delta}_i),  a\big) \Big)  \\  
    &\st \quad \|\boldsymbol{\delta}_i \|_{\infty} \leq \epsilon, \;\forall \: i = 1, \dots, n.
\end{split}
\end{equation}
\normalsize
In \eqref{eq: loss_abstain}, tuning $\lambda_1$ and $\lambda_2$ controls the trade-off between standard and robust accuracy. Furthermore, to obtain non-trivial  results,  IBP-relaxation should be incorporated during training for  the minimization sub-problems in $L_\textrm{robust}$ and $L_\textrm{robust}^{\textrm{abstain}} $ \citep{sheikholeslamiprovably,gowal2018effectiveness}. 
 
\vspace{-2mm}
\section{Verification of Neural Networks with Multiple Detection Classes}
\begin{figure}
\vspace{-2mm}
\centerline{\includegraphics[width=1\columnwidth]{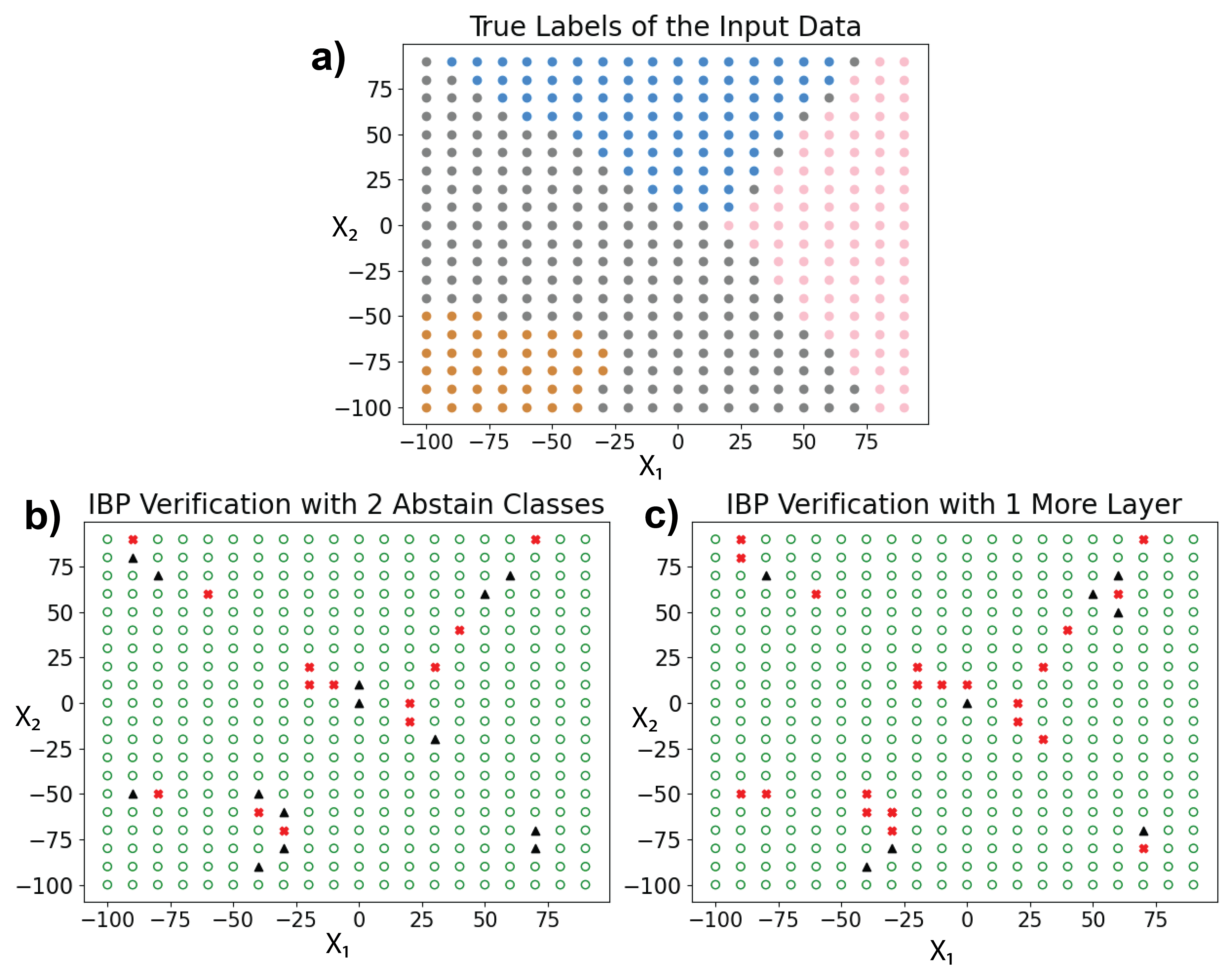}}
\vspace{-3mm}
\caption{\footnotesize The IBP verification for $400$ input data points of $2$-layer and $3$-layer neural networks. Part \textbf{(a)} shows the assigned four labels to  data points. Part \textbf{(b)}  demonstrates that IBP can verify $14$ points using one of two abstain classes (black triangles), while it cannot verify $13$ data points (red $\times$). \textbf{c)} shows that when IBP is applied to a network with one more layer and one detection class, $8$ points are verified by the detection class, while it fails to verify $21$ points. 
The description of both networks can be found in Appendix~\ref{appendix: network_structure}. }
\label{fig: ibp}
\end{figure}

\vspace{-2mm}
\label{sec:verification}
\textbf{Motivation:}  The set of all adversarial images that can be generated within the $\epsilon$-neighborhood of clean images might not be detectable only by a single detection class. Hence, the robust verified accuracy of the joint classifier and detector can be enhanced by introducing multiple abstain classes instead of a single abstain class to detect adversarial examples.
This observation is illustrated in a simple example in Appendix~\ref{appendix: multi_abstain_motivation} where we theoretically show that $2$ detection classes can drastically increase the performance of the detector compared to $1$ detection class. 
Note that a network with multiple detection classes can be equivalently modeled by another network with one more layer and a single abstain class. This added layer, which can be a fully connected layer with a max activation function, can merge all abstain classes and collapse them into a single  class. \ul{\it Thus, any $L$-layer neural network with multiple abstain classes can be equivalently modeled by an $L+1$-layer neural network with a single abstain class.} 
However, the  performance of verifiers such as IBP reduces as we increase the number of layers. The reason is that increasing the number of layers leads to looser bounds in~\eqref{eq:IBP} for the last layer. To illustrate this fact, Figure~\ref{fig: ibp} shows that the number of verified points by a $2-$layer neural network is higher than the number of points verified by an equivalent network with  $3$ layers. 


Thus, it is beneficial to train/verify the original $L$-layer neural network with multiple abstain classes instead of $L+1$-layer network with a single abstain class. This fact will be illustrated further in the experiments on MNIST and CIFAR-10 datasets depicted in Figure~\ref{fig: shallow}. Next, we present how one can verify a network with multiple abstain classes:

Let $a_1, a_2, \dots, a_M$ be $M$ abstain classes detecting adversarial samples. A sample is considered  adversarial  if the network's output is any of the $M$ abstain classes.  A neural network with $K$ regular classes and $M$ abstain classes outputs the label of a given sample as
$
    \hat{y}(\bx) = \argmax_{i \in \{1, \dots, K, a_1, \dots, a_M\}} [\bz_L(\bx)]_i.
$
An input $(\mathbf{x},y)$ is verified if the network either correctly classifies it as class $y$ or assigns it to any of the explicit $M$ abstain classes. More formally and following equation~\eqref{eq: verification_original_problem1}, the neural network is verified for input~$\bx_0$ against a target class $k$ if 
\begin{equation}
 \label{eq: verification_min_max_multi_abstain}
0 \leq \min_{\bz_L \in \mathcal{Z}(\bx_0, \epsilon)} \max \left\{\bc_{yk} ^T \bz_{L},  \bc_{a_1 k} ^T \bz_{L}, \dots, \bc_{a_M k} ^T \bz_{L}\right\}. 
\end{equation}
Since the set~$\mathcal{Z}(\bx_0, \epsilon)$ is nonconvex, verifying~\eqref{eq: verification_min_max_multi_abstain} is computationally expensive. The next subsections present two verifiers for solving~\eqref{eq: verification_min_max_multi_abstain} based on IBP and $\beta$-crown. 

\vspace{-3mm}
\subsection{Verification with IBP}
\vspace{-2mm}
Using the IBP mechanism to relax the non-convex set~$\mathcal{Z}(\bx_0, \epsilon)$ leads to the following result for a given network with $M$ detection classes:
\vspace{-2mm}
\begin{theorem}
\label{thm: verification_problem_lower_bound}
Condition~\eqref{eq: verification_min_max_multi_abstain} is satisfied if for all $k \neq y$:
\vspace{-2mm}
\begin{align} 
\small
\min_{\boldsymbol{\eta} \in \mathcal{P}}   \quad \max_{\underline{\bz}_{L-1} \leq \bz_{L-1} \leq \bar{\bz}_{L-1}}   - c_k(\boldsymbol{\eta})^T (\bW_L \bz_{L-1} + \bb_L)
\label{eq: verification_min_max_original}
\end{align}
is greater than or equal to zero where
$
\mathcal{P} = \{(\eta_0, \dots, \eta_M) | \sum_{i=0}^{M} \eta_i = 1, \eta_i \geq 0,  \forall i=0,1, \ldots, M \},    
$ and $c_k(\boldsymbol{\eta}) = \eta_0 \bc_{yk} + \eta_1 \bc_{a_1k}  \dots + \eta_M \bc_{a_Mk}.$ Here, the bounds $\underline{\bz}_{L-1}$ and $\bar{\bz}_{L-1}$ are obtained according to~\eqref{eq:IBP}.
\end{theorem}
Unlike \eqref{eq: verification_min_max_multi_abstain}, the condition in \eqref{eq: verification_min_max_original} is easy to verify computationally. To understand this, let us define 
\begin{equation} \label{eq: J-definition}
    J_k(\boldsymbol{\eta}) = \max_{\underline{\bz}_{L-1} \leq \bz_{L-1} \leq \bar{\bz}_{L-1}} - c_k(\boldsymbol{\eta})^T (W_L \bz_{L-1} + \bb_L).
\end{equation} 
Then, our aim in \eqref{eq: verification_min_max_original} is to minimize $J_k (\boldsymbol{\eta})$ over $\mathcal{P}$. 

First, notice that the maximization problem~\eqref{eq: J-definition} can be solved in closed form  as described in Step~\ref{Step:updateZ} of Algorithm~\ref{alg: bregman divergence}. Consequently, one can rely on Danskin's Theorem~\citep{danskin2012theory} to compute the subgradient of the function $J_k(\cdot)$. 
Thus, to minimize $J_k(\cdot)$ in \eqref{eq: verification_min_max_original}, we can rely on the Bregman proximal (sub)gradient method (see~\citep{gutman2018unified} and the references therein). This algorithm is guaranteed to find $\epsilon-$ accurate solution to \eqref{eq: verification_min_max_original} in $T = O(1/\sqrt{\epsilon})$ iterations--see~\citep[Corollary 2]{gutman2018unified}.

\begin{algorithm}
    \caption{IBP verification of the network with multiple detection classes against class $k$}
    \label{alg: bregman divergence}
	\begin{algorithmic}[1]
	 \STATE \textbf{Parameters}: Stepsize~$\nu>0$, number of iterations~$T$.
	 \STATE Initialize ${\eta}_0 = 1$ and $  {\eta}_1  = \ldots = {\eta}_M= 0.$
	  \FOR {$t = 0, 1, \ldots, T$}
	    \STATE\label{Step:updateZ} 
	   $[\bz^{*}_{L-1}]_j = \begin{cases*}
      [\underline{\bz}_{L-1}]_j & if $[\bW_L^T c_k(\boldsymbol{\eta}^t)]_j \geq 0 $ \\
      [\bar{\bz}_{L-1}]_j        & otherwise.
    \end{cases*}.$
    \STATE Set $\eta_m^{t+1} = \frac{\eta_m^{t} \exp(-2\nu  (\bz_{L-1}^{*t})^T \bW_{L}^T \bc_{a_m k}   )}{\sum_{j=0}^{M} \eta_j^{t} \exp(-2\nu (\bz_{L-1}^{*t})^T \bW_{L}^T \bc_{a_j k}    )}$, for all  $\quad \quad m \in [M]$ where $a_0 = y$.
    	\ENDFOR
	\end{algorithmic}
\end{algorithm}
\vspace{-0.2cm}
\begin{remark} \textbf{Time Complexity Comparison:} \citet{sheikholeslamiprovably} finds $J_k (\boldsymbol{\eta})$ using sorting and comparing a finite number of calculated values based on the last two layer weight matrices. While their algorithm finds the \textbf{exact} solution of $\eta$, it is limited to the one-dimensional case $(M=1)$. In this scenario, their algorithm requires $\mathcal{O}(n_{L-1} \log(n_{L-1}))$ evaluations to find the optimal $\eta$ where $n_{L-1}$ represents the number of nodes in the one to the last layer. Alternatively, our algorithm gives an $\epsilon$-optimal solution in order of $\mathcal{O}(\frac{1}{\epsilon})$. By choosing $\epsilon = \mathcal{O}(\frac{1}{n_{L-1}})$, Algorithm~\ref{alg: bregman divergence} has almost the same  complexity as Algorithm~1 in~\citet{sheikholeslamiprovably} in the one dimensional case. When $M > 1$, their algorithm cannot be utilized, while our algorithm finds the solution with the same order complexity. Note that the order complexity, in this case, is linear with respect to $M$.  
\end{remark}
\vspace{-3mm}

\subsection{Verification with \texorpdfstring{$\beta$}--Crown}
\vspace{-0.3cm}
IBP verification is computationally efficient but less accurate than $\beta$--Crown. Hence, we focus on $\beta$--Crown verification of networks with multiple abstain classes in this section to obtain a tighter verifier. To this end, we will find a sufficient condition for~\eqref{eq: verification_min_max_multi_abstain} using the lower-bound technique of~\eqref{eq: beta_crown_subproblem} in $\beta$--Crown. In particular, by switching the minimization and maximization in \eqref{eq: verification_min_max_multi_abstain} and using the $\beta$--Crown lower bound~\eqref{eq: beta_crown_subproblem}, we can find a lower-bound of the form 
\begin{align}
\small
\label{eq: beta_crown_subproblem_multiple_abstain}
    & \min_{\bz_L \in \mathcal{Z}(\bx_0, \epsilon)}  \max \{\bc_{yk} ^T \bz_{L},  \bc_{a_1 k} ^T \bz_{L}, \dots, \bc_{a_M k} ^T \bz_{L}\} 
    \geq  \nonumber \\ & \max_{\boldsymbol{\eta} \in \mathcal{P}, \boldsymbol{\alpha}, \boldsymbol{\beta} \geq 0} 
    G(\bx_0, \boldsymbol{\alpha}, \boldsymbol{\beta} , \boldsymbol{\eta} ).
\end{align}
The details of this inequality and the exact definition of the function $G(\cdot)$ is provided in Appendix~\ref{appendix: derivation_G_inequality}. Note that any feasible solution to the right-hand side of~\eqref{eq: beta_crown_subproblem_multiple_abstain} is a valid lower bound to the original verification problem (left-hand-side). Thus, in order for~\eqref{eq: verification_min_max_multi_abstain} to be satisfied, it suffices to find a feasible $(\boldsymbol{\alpha}, \boldsymbol{\beta} , \boldsymbol{\eta})$ such that $G(\bx_0, \boldsymbol{\alpha}, \boldsymbol{\beta} , \boldsymbol{\eta} ) \geq 0$. To optimize the RHS of~\eqref{eq: beta_crown_subproblem_multiple_abstain} in Algorithm~\ref{alg: beta_crown_multiple_abstain}, we utilize AutoLirpa library of~\citep{zhang2019towards} for updating $\boldsymbol{\alpha}$, and use Bregman proximal subgradient method to update $\boldsymbol{\beta}$ and $\boldsymbol{\eta}$ -- See appendix~\ref{appendix: bregman_divergence}. We use Euclidean norm Bregman divergence for updating $\boldsymbol{\beta}$ and Shannon entropy Bregman divergence for $\boldsymbol{\eta}$ to obtain closed-form updates. 
\begin{algorithm}
    \caption{$\beta$--Crown verification of networks with multiple detection classes}
    \label{alg: beta_crown_multiple_abstain}
	\begin{algorithmic}[1]
     \STATE \textbf{Input}: number of iterations $T$, number of iterations in the inner-loop $T_0$, Step-size $\gamma$. 
	    \FOR {$t = 0, 1, \ldots, T$}
            \STATE Update $\alpha$ using AutoLirpa ~\citep{zhang2019towards}
            \FOR {$k = 0, 1, \ldots, T_0$}
                \STATE $\boldsymbol{\beta} = [\boldsymbol{\beta} + \gamma \frac{\partial G(\bx_0, \boldsymbol{\alpha}, \boldsymbol{\beta} , \boldsymbol{\eta} }{\partial \boldsymbol{\beta}}]_{+}$,
                \STATE $\eta_m^{\textrm{new}} = \frac{\eta_m^{\textrm{old}} \exp(2\gamma \frac{\partial   G(\bx_0, \boldsymbol{\alpha}, \boldsymbol{\beta} , \boldsymbol{\eta})    }{\partial \eta_m})}{\sum_{j=0}^{M} \eta_j^{\textrm{old}} \exp(2\gamma \frac{\partial  G(\bx_0, \boldsymbol{\alpha}, \boldsymbol{\beta} , \boldsymbol{\eta})  }{\partial \eta_j})} \: \: \forall m \in [M].$
            \ENDFOR
		\ENDFOR
		\end{algorithmic}
\end{algorithm}

$[w]_+ = \max(w, 0)$ denotes the projection to the non-negative orthant in Algorithm~\ref{alg: beta_crown_multiple_abstain}.

\vspace{-3mm}
\section{Training of Neural Networks with Multiple Detection Classes}
\label{sec:training}
\vspace{-0.3cm}
We follow a similar combination of loss functions to train a neural network consisting of multiple abstain classes as in~\eqref{eq: loss_abstain}. While the last term ($L_{\textrm{Natural}}$) can be computed efficiently, the first and second terms cannot be computed efficiently because even evaluating the functions~$ L_{\textrm{Robust}}$ and $L_{\textrm{Robust}}^{\textrm{Abstain}}$ requires maximizing nonconcave functions. Thus, we will minimize their upper bounds instead of minimizing these two terms. Particularly, following~\citep[Equation~(17)]{sheikholeslami2020minimum}, we use $\bar{L}_{\textrm{Robust}}$ as an upper-bound to $L_{\textrm{Robust}}$. This upper-bound is obtained by the IBP relaxation procedure described in subsection~\ref{subsec:IBP}. To obtain an upper-bound for the Robust-Abstain loss term $L_{\textrm{Robust}}^{\textrm{Abstain}}$ in \eqref{eq: loss_abstain}, let us first start  by clarifying its definition in the multi-abstain class scenario:
\vspace{-2mm}
\begin{align}
\small    
L_{\textrm{Robust}}^{\textrm{Abstain}} = \max_{\delta_1, \dots, \delta_n} \: \frac{1}{n} \sum_{i=1}^n  &\min \Big\{  \ell_{\textrm{xent}} \big(\bz_L(\bx_i + \delta_i),  y_i \big), \nonumber \\ 
&\min_{m \in [M]} \ell_{\textrm{xent}} \big(\bz_L(\bx_i + \delta_i),  a_m \big)  \Big\}.
\label{eq: multi_abstain_loss1}
\end{align}
This definition implies that  the classification is considered ``correct'' for a given input if the predicted label is the ground-truth label or if it is assigned to one of the abstain classes. Since the maximization problem w.r.t. $\{\delta_i\}$ is nonconcave, it is hard to even evaluate $L_{\textrm{Robust}}^{\textrm{Abstain}}$. Thus, we minimize an efficiently computable upper bound of this loss function as described in Theorem~\ref{thm: abstain_loss}. 

\begin{theorem}
\label{thm: abstain_loss}
Let $\ell_{\textrm{Robust}}^{\textrm{Abstain}} (\bx,y)$ is defined as:
\begin{align*}
\small
 \max_{\|\delta\| \leq \epsilon} \min \Big\{  \ell_{\textrm{xent}} \big(\bz_L(\bx + \delta),  y \big), \min_{m \in [M]} \ell_{\textrm{xent}} \big(\bz_L(\bx + \delta),  a_m \big) \Big\}    
\end{align*}
Then,
\begin{equation}
\ell_{\textrm{Robust}}^{\textrm{Abstain}}(\bx, y) \leq \bar{\ell}_{\textrm{Robust}}^{\textrm{Abstain}}(\bx, y) = \ell_{\textrm{xent} \setminus \mathcal{A}_0} (J(\bx), y),      
\end{equation}
where $J(\bx)$ is a vector whose $k$-th component equals $J_k(\bx)$ as defined in~\eqref{eq: J-definition} and $\ell_{\textrm{xent} \setminus \mathcal{A}_0}(\bx_0, y) := -\log \Bigg(\frac{\exp(\mathbf{e}_y^T\bz_L(\bx_0))}{\sum_{i \in \mathcal{I} \setminus \mathcal{A}_0} \exp(\mathbf{e}_i^T\bz_L(\bx_0))} \Bigg)$. Here,  $\mathcal{I} = \{1, \dots, K, a_1, \dots, a_M\}$ is the set of all classes (true labels and abstain classes) and $\mathcal{A}_0 = \{a_1, \dots, a_M\}$ is the set of abstain classes.
\end{theorem}
Notice that the definition of~$\ell_{\textrm{xent} \setminus \mathcal{A}_0}(\bx_0, y)$ removes the terms corresponding to the abstain classes in the denominator. 
This definition is less restrictive toward abstain classes compared to incorrect classes. Thus, for a given sample, it is more advantageous for the network to classify it as an abstain class instead of an incorrect classification. This mechanism enhances the performance of the network in detecting adversarial examples by abstain classes, while it does not have an adverse effect on the performance of the network on natural samples. Note that during the evaluation/test phase, this loss function does not change the final prediction of the network for a given input since the winner (the entry with the highest score) remains the same. Overall, we upper-bound the loss  in~\eqref{eq: loss_abstain} by replacing $L_{\textrm{Robust}}$ with the IBP relaxation approach utilized in~\citet{gowal2018effectiveness, sheikholeslamiprovably} and replacing $L_{\textrm{Robust}}^{\textrm{Abstain}}$ with $\bar{L}_{\textrm{Robust}}^{\textrm{Abstain}} = \frac{1}{n} \sum_{i=1}^n \bar{\ell}_{\textrm{Robust}}^{\textrm{Abstain}}(\bx_i, y_i)$ presented in Theorem~\ref{thm: abstain_loss}. Thus our total training loss can be presented as:
\begin{equation}
\label{eq: final_loss}
L_{\textrm{Total}} = \bar{L}_{\textrm{Robust}} + \lambda_1 \bar{L}_{\textrm{Robust}}^{\textrm{Abstain}}  +  \lambda_2 L_{\textrm{Natural}}
\end{equation}
Algorithm~\ref{alg: training} describes the procedure of optimizing~\eqref{eq: final_loss} on a joint classifier and detector with multiple abstain classes. 

\begin{algorithm}
    \caption{Robust Neural Network Training}
    \label{alg: training}
	\begin{algorithmic}[1]
	 \STATE \textbf{Input}: Batches of data $\mathcal{D}_1, \dots, \mathcal{D}_R$, step-size $\nu$, $\btheta(L)$: optimization parameters for loss $L$. 
	    \FOR {$t = 1, \ldots, R$}
        \STATE Compute $J_o(\bx) \: \forall \: \bx \in \mathcal{D}_t, \: \forall o \in [K]$ by Algorithm~\ref{alg: bregman divergence}.
        
        \STATE Compute $\bar{L}_{\textrm{Robust}}$ on Batch $\mathcal{D}_t$ as~\citet{gowal2018effectiveness}.
        
        \STATE Compute $\bar{L}_{\textrm{Robust}}^{\textrm{abstain}}$ on Batch $\mathcal{D}_t$ using Theorem~\ref{thm: abstain_loss}.
        
        \STATE 
        \vspace{-5mm}
        \begin{align*}
            \small \btheta(L) = \btheta(L) - \nu \nabla_{\btheta} \big(\bar{L}_{\textrm{Robust}} + \lambda_1 \bar{L}_{\textrm{Robust}}^{\textrm{abstain}} + \lambda_2 L_{\textrm{Natural}}\big)
        \end{align*}
		\ENDFOR
		\end{algorithmic}
\end{algorithm}

\subsection{Addressing model degeneracy}
\vspace{-3mm}
Having multiple abstain classes can potentially increase the capacity of our classifier to detect adversarial examples. However, as we will see in Figure~\ref{fig: bar_chart} (10 abstains, unregularized), several abstain classes collapse together and  capture similar adversarial patterns. Such a phenomenon, referred to as ``model degeneracy'' and illustrated with an example in Appendix~\ref{appendix: multi_abstain_motivation} will prevent us from fully utilizing all abstain classes. 
\begin{figure*}[ht]
\begin{center}
\vspace{-2mm}
\centerline{\includegraphics[width=1\textwidth]{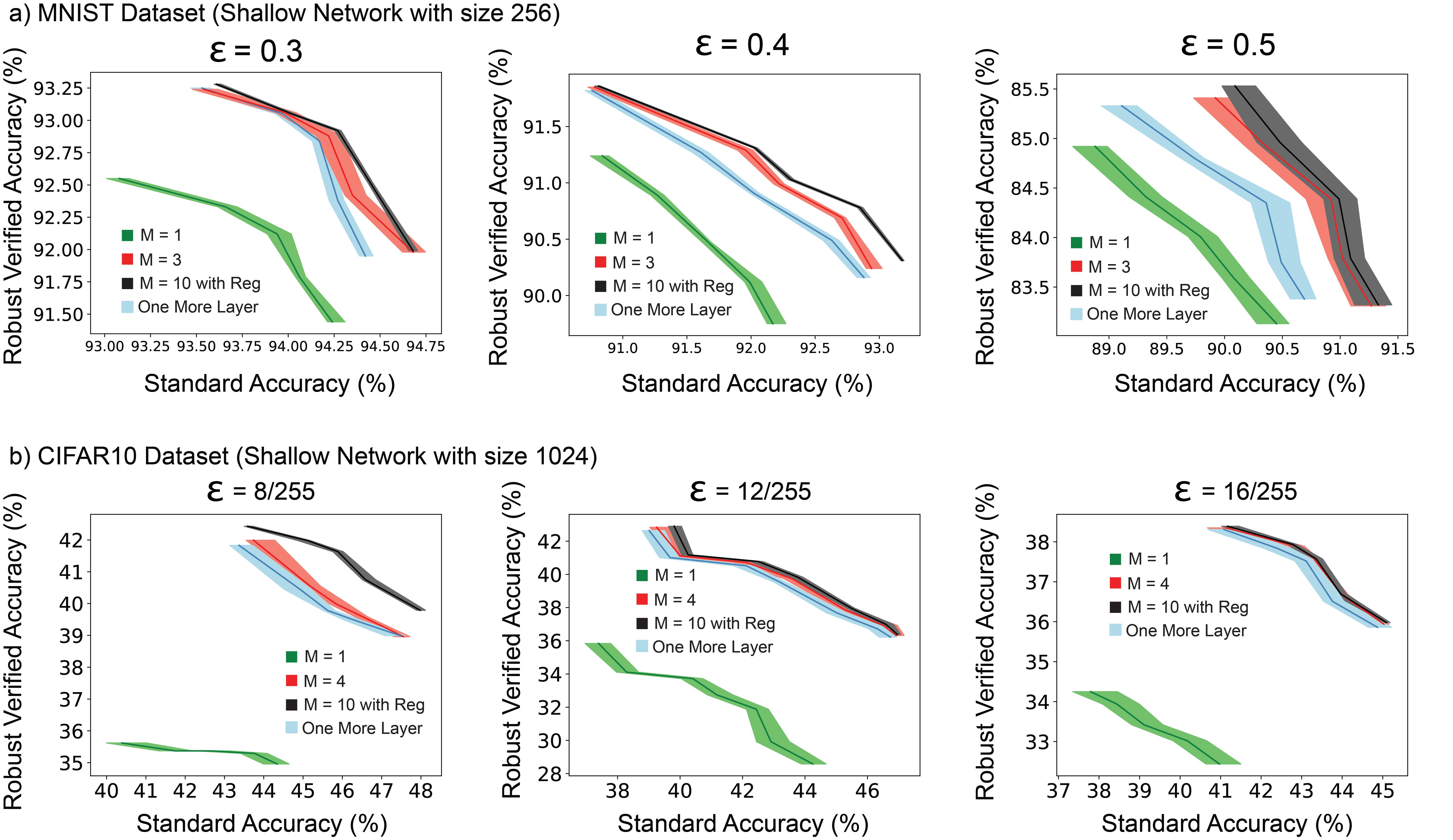}}
\caption{\footnotesize Performance of Multiple-abstain shallow networks on MNIST and CIFAR-10 datasets. We compared multiple abstain neural networks (both regularized and non-regularized versions) with the single abstain networks and networks with one more layer. The above and below rows demonstrate the trade-off between standard and robust verified accuracy on MNIST and CIFAR-10 datasets.}
\vspace{-4mm}
\label{fig: shallow}
\end{center}
\end{figure*}

To address this issue,  we impose a regularization term to the loss function such that the network utilizes all abstain classes in balance. We aim to make sure the $\boldsymbol{\eta}$ values are distributed nearly uniformly, and there are no \textit{idle} abstain classes. Let $\boldsymbol{\eta}^{ik}$, $\bz_{L-1}(\bx_i)$, and $y_i$ be the abstain vector corresponding to the sample $\bx_i$ verifying against the target class $k$, the output of the layer $L-1$, and the assigned label to the data point $\bx_i$ respectively. Therefore, the regularized verification problem over $n$ given samples takes the following form:
\begin{align}
\label{eq: verification_regularized}
\small
& \min_{\boldsymbol{\eta}^1, \dots, \boldsymbol{\eta}^n \in \mathcal{P}} \;  \sum_{i=1}^{n} \sum_{k\neq y_i}   \max_{\underline{\bz}(\bx_i) \leq \bz_{L-1} \leq \bar{\bz}(\bx_i)} - c_k(\boldsymbol{\eta}^{ik})^T f(\bz_{L-1})  \nonumber \\ 
 &+ \mu \| \Big[ \frac{\gamma \mathbf{1}}{M+1} - \frac{\sum_{j=1}^n \sum_{o\neq y_i} \boldsymbol{\eta}^{jo}}{n(K-1)} \Big]_+ \|^2,
\end{align}

\vspace{-2mm}
where $f(\bz_{L-1}) = \bW_L \bz_{L-1} + \bb_L$. The above regularizer penalizes the objective function if the average value of $\boldsymbol{\eta}$ coefficient corresponding to a given abstain class over all batch samples is smaller than a threshold (which is determined by the hyper-parameter $\gamma$). In other words, if an abstain class is not contributing enough to detect adversarial samples, it will be penalized accordingly. Note that if $\gamma$ is larger, we penalize an \textit{idle} abstain class more. 

Note that in the unregularized case, the optimization of parameters $\boldsymbol{\eta}^{ik}$ are independent of each other. In contrast, by adding the regularizer described in~\eqref{eq: verification_regularized}, we require to optimize $\boldsymbol{\eta}^{ik}$ of different samples and target classes jointly as they are coupled in the regularization term. Since optimizing~\eqref{eq: verification_regularized} over the set of all $n$ data points is infeasible for datasets with many samples, we solve the problem by choosing small data batches ($\leq 64$). We utilize the same Bregman divergence procedure used in Algorithm~\ref{alg: bregman divergence}, while the gradient with respect to $\boldsymbol{\eta}^{ik}$ takes the regularization term into account as well.

\noindent\textbf{Hyper-parameter tuning compared to the single-abstain scenario.} In contrast to~\cite{sheikholeslamiprovably}, our methodology has the additional hyper-parameter $M$ (the number of abstain classes). Tuning this hyper-parameter can be costly since, in each run, we have to change the architecture (and potentially the stepsizes of the algorithm). This can be viewed as a potential additional computational overhead for training. However, as we observed in our experiments, setting $M=K$ combined with our regularization mechanism always leads to significant performance gains over training with a single abstain class. Thus, tuning the hyper-parameter $M$ is unnecessary when we apply the regularization mechanism. In that case, we have two other hyper-parameters: $\gamma$ is chosen from the set $\{0.1, 0.2, 0.5, 1, 2, 5, 10\}$ over $K+M$. The experiments show that $1$ consistently works the best across different datasets and architectures. Further, $\mu$ in~\eqref{eq: verification_regularized} is the regularization mechanism hyper-parameter. We determine $\mu$ by cross-validation over a wide range of numbers in $[0.1, 10]$. The experiments show that the optimal value of $\mu$ is between $[0.8, 1.5]$ depending on the network, dataset, and $\epsilon$.

\vspace{-3mm}
\section{Numerical results}
\label{sec: numerical_results}
\vspace{-4mm}
We devise diverse experiments on shallow and deep networks to investigate the effectiveness of joint classifiers and detectors with multiple abstain classes.

\vspace{-3mm}
\subsection{Training Setup}
\vspace{-2mm}
To train the networks on MNIST and CIFAR-10 datasets, we use Algorithm~\ref{alg: training} as a part of an optimizer scheduler. Initially, we set $\lambda_1 = \lambda_2 =0$. Thus, the network is trained without considering any abstain classes. In the second phase, we optimize the objective function~\eqref{eq: final_loss}, where we linearly increase $\epsilon$ from $0$ to $\epsilon_{\textrm{train}}$. Finally, we further tune the network on the fixed $\epsilon=\epsilon_{\textrm{train}}$. On both MNIST and CIFAR-10 datasets, we have used an Adam optimizer with a learning rate $5 \times 10^{-4}$. The networks are trained with four NVIDIA V100 GPUs. 
The trade-off between standard accuracy on clean images and robust verified accuracy can be tuned by changing $\lambda_2$ from $0$ to $+\infty$ where the larger values correspond to more robust networks. For the networks with the regularizer addressing the model degeneracy issue, we choose $\gamma$ by tuning it in the $[\frac{0.1}{K+M}, \frac{1.5}{K+M}]$. Our observations on both MNIST and CIFAR-10 datasets for different $\epsilon$ values show that the optimal value for $\gamma$ is consistently close to $\frac{1}{K+M}$. Thus, we suggest choosing hyper-parameter $\gamma = \frac{1}{K+M}$ where $K$ is the number of labels, and $M$ is the number of detection classes. The optimal value for $M$ is $4$ for the CIFAR-10 and $M=3$ for the MNIST dataset. By adding the "model degeneracy" regularizer, the obtained network has nearly the same performance for $M \in [4, 2K]$. Overall, we suggest to choose $M = K$ and $\gamma = \frac{1}{K+M}$ as the default values for hyper-parameters $M$ and $\gamma$.  

The model's robust verified accuracy will generally be increased by changing $\lambda_2$ from $0$ to large values. As compensation, the standard accuracy of the model is reduced. Therefore, $\lambda_2$ determines the trade-offs between the standard and robust verified accuracy. The tradeoff curves presented in the figures are obtained by changing $\lambda$ from $0$ to $100$. 
\begin{figure}
\begin{center}
\centerline{\includegraphics[width=1\columnwidth]{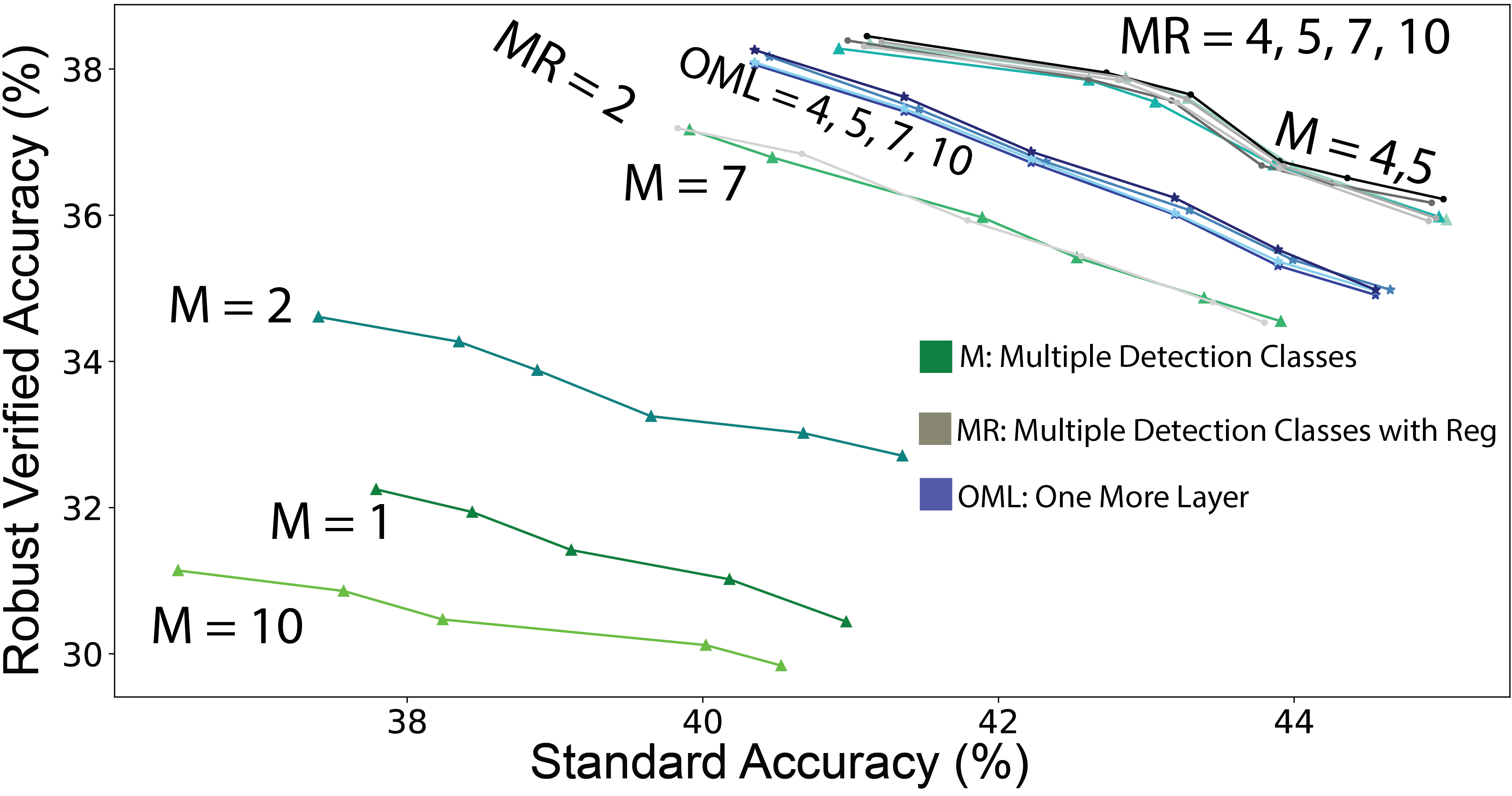}}
\caption{\footnotesize Performance of shallow networks with multiple detection classes without regularization, with regularization, and the network with one more layer on CIFAR-10 dataset. As demonstrated, the networks with regularization work consistently well for $4 \leq M\leq 10$, which is very close to the best performance we get from networks without regularization with $4$ detection classes.}
\label{fig: shallow_different_M}
\vspace{-0.8cm}
\end{center}
\end{figure}

\subsection{Robust Verified Accuracy on Shallow Networks}
\vspace{-2mm}
In the first set of experiments depicted in Figure~\ref{fig: shallow}, we compare the performance of shallow networks with the fully optimized number of abstain classes to the single abstain network, the network with an additional layer, and the network with $M=K$ regularized by Equation~\eqref{eq: verification_regularized}. The shallow networks have one convolutional layer with sizes $256$ and $1024$ for MNIST and CIFAR-10 datasets, respectively. This convolutional layer is connected to the second (last) layer consisting of $K+M$ (20 for both MNIST and CIFAR-10) nodes. The optimal number of abstain classes is obtained by changing the number of them from $1$ to $20$ on both MNIST and CIFAR-10. The optimal value for the network trained on MNIST is $M=3$, and $M=4$ for the CIFAR-10 dataset.
\begin{figure}[ht]
  \begin{center}
    \includegraphics[width=1\columnwidth]{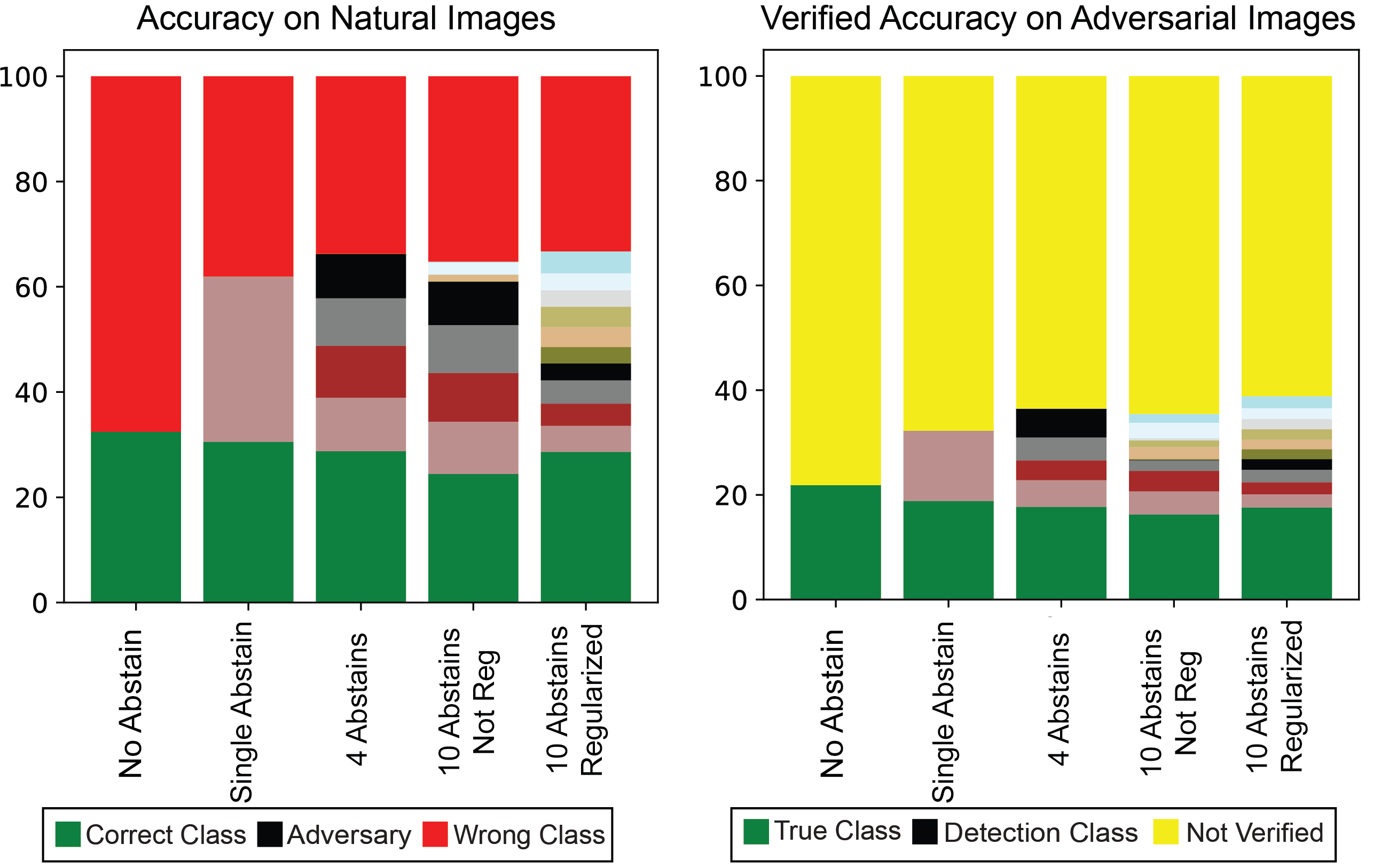}
    \vspace{-4mm}
    \caption{\footnotesize Distribution of Natural and adversarial images over different abstain classes on CIFAR-10 dataset. When there are 10 abstain classes, model degeneracy leads to lower performance than the baseline. Adding the regularization term (rightmost column)  will utilize all abstain classes and enhance standard and robust verified accuracy. Standard accuracy is the proportion of correctly classified natural images, while robust verified accuracy is the proportion of images that are robust against all adversarial attacks within the $\epsilon$-neighborhood ($\epsilon = \frac{16}{255}$).}
    \vspace{-4mm}
\label{fig: bar_chart}
  \end{center}
\end{figure}
Moreover, we compare the optimal multi-abstain shallow network to two other baselines: One is the network with the number of abstain classes equal to the number of regular classes ($M=K$) trained via the regularizer described in~\eqref{eq: verification_regularized}. The other is a network with one extra layer than the shallow network. This network has $K+M$ nodes in one to the last layer and $K+1$ nodes in the last layer compared to the shallow network. Ideally, the set of models can be supported by such a network is a super-set of the original shallow network. Therefore, it has more capacity to classify images and detect adversarial attacks. However, due to the training procedure of IBP, which is sensitive to a higher number of layers (the higher the number of layers, the looser, the lower and upper bounds), we obtain better results with the original network with multiple abstain classes. 
\begin{figure*}[ht]
\begin{center}
\centerline{\includegraphics[width=1\textwidth]{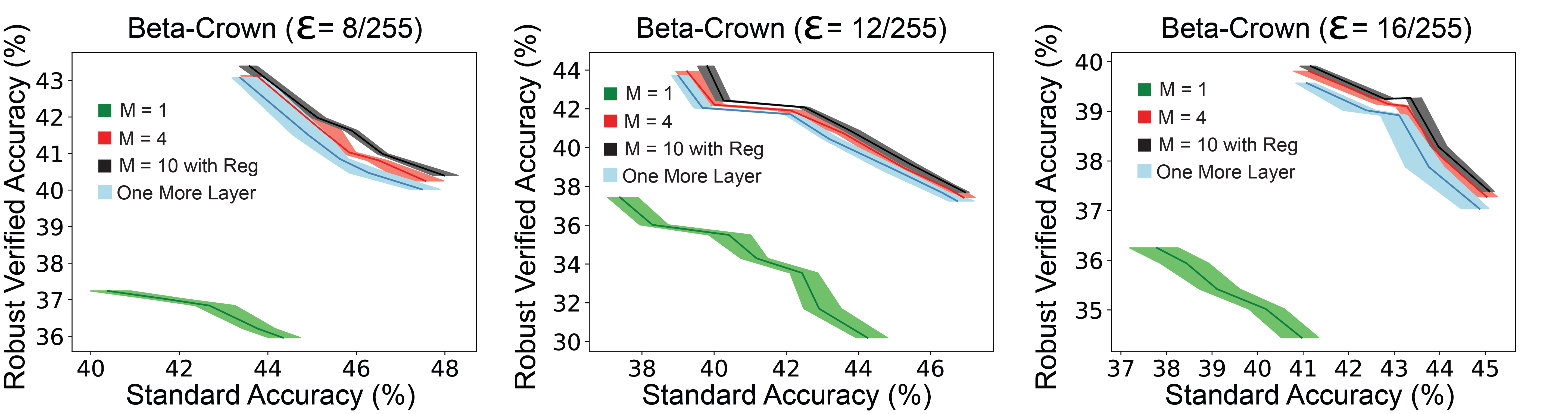}}
\vspace{-3mm}
\caption{\footnotesize Performance of $\beta$-crown on verification of Neural Networks with single abstain, $4$ abstain classes, $10$ abstain classes with regularized, and networks with one more layer (single abstain) on CIFAR-10 dataset. $M=1$ coincides with~\cite{sheikholeslamiprovably}.}
\label{fig: beta}
\end{center}
\end{figure*}

Next, in Figure~\ref{fig: shallow_different_M}, we investigate the effect of changing the number of abstain classes of the shallow  network described above. We observe that the unregularized network and the network with one more layer are much more sensitive to the change of $M$ than the regularized version. This means we can use the regularized network with the same performance while it does not require to be tuned for the optimal $M$. In the unregularized version, by increasing the number of abstain classes from $M=1$ to $M=4$, we see improvement. However, after this threshold, the network performance drops gradually such that for $M=10$ where the number of labels and abstain classes are equal ($M = K = 10$), the performance of the network, in this case, is even worse than the single-abstain network due to the model degeneracy of the multi-abstain network. However, the network trained on the regularized loss maintains its performance when $M$ changes from the optimal value to larger values. 

Figure~\ref{fig: bar_chart} shows the percentage of adversarial examples captured by each abstain class ($M=10$) on CIFAR-10 for both regularized and non-regularized networks. The hyper-parameter $\gamma$ is set to $\frac{1}{K+M} = 0.05$. Next, we illustrate the performance of networks trained in the first set of experiments by $\beta$-crown in Figure~\ref{fig: beta}. The networks verified by Beta-crown have $1\%$ to $2\%$ improvement in robust accuracy compared to the same networks verified by IBP.

\vspace{-2mm}
\subsection{Performance on Deep Neural Networks}
\vspace{-2mm}
Networks with multiple abstain classes achieve a superior trade-off between standard and verified robust accuracy on the deep networks as well. To demonstrate the performance of multiple abstain classes compared to state-of-the-art approaches, we trained deep neural networks on MNIST and CIFAR-10 datasets with different $\epsilon$ values. The results are reported in Table~\ref{tab:deep-network}. The structure of the trained deep network is the same as the one described in~\citet{sheikholeslamiprovably} (see Appendix~\ref{appendix: implementation_details}). 

While the verified robust accuracy guarantees the robustness of the networks against \textbf{all} attacks within the $\epsilon$-neighborhood of each given test data point, one can argue that in many practical situations, being robust against certain adversarial attacks such as PGD attack~\citep{madry2017towards} is sufficient. Table~\ref{tab:pgd-mnist} and Table~\ref{tab:pgd-cifar10} demonstrate the performance of several state-of-the-art approaches for training robust neural networks against adversarial attacks on the MNIST and CIFAR-10 datasets, respectively. The performances are evaluated by the standard accuracy and robustness against the PGD attack on the test samples. The chosen $\epsilon$ is $0.4$ for MNIST and $8/255$ for CIFAR-10, and the attacks are applied to each test sample using $100$ iterations of the projected gradient descent (PGD). Our method achieves the best robustness against PGD attacks on both MNIST and CIFAR-10 datasets.

\begin{table}[]
\begin{tabular}{|c|c|c|}
\hline
\textbf{Method}     & \textbf{St Err} & \textbf{PGD Success} \\ \hline
\citep{madry2017towards}               & $2.91\%$                    & $11.35\%$                       \\ \hline
IBP                 & $2.27\%$                    & $6.68\%$                        \\ \hline
IBP Crown  & $2.17\%$                    & $12.06\%$                       \\ \hline
\small \citep{balunovic2019adversarial} & $2.77\%$                    & $15.31\%$                       \\ \hline
\citep{sheikholeslamiprovably}      & $4.74\%$                    & $4.15\%$                        \\ \hline
\citep{aquino2022robustness}         & $4.81\%$                    & $5.86\%$               \\ \hline
Our Method        & $4.97\%$                    & $\mathbf{3.91\%}$                        \\ \hline
\end{tabular}
\vspace{-0.4cm}
\caption{\label{tab:pgd-mnist}Standard Error and PGD attack success rate on the MNIST dataset for different state-of-the-art approaches. The chosen $\epsilon$ for the PGD attack equals $0.4$.}
\end{table}
\begin{table}[]
\begin{tabular}{|c|c|c|}
\hline
\textbf{Method}     & \textbf{St Err} & \textbf{PGD Success} \\ \hline
\citep{madry2017towards}               & $49.78\%$                    & $68.48\%$                       \\ \hline
IBP                 & $50.51\%$                    & $65.23\%$                        \\ \hline
IBP Crown  & $54.02\%$                    & $65.42\%$                       \\ \hline
\small \citep{balunovic2019adversarial} & $48.3\%$                    & $69.81\%$                       \\ \hline
\citep{sheikholeslamiprovably}      & $55.60\%$                    & $63.63\%$                        \\ \hline
\citep{aquino2022robustness}         & $50.25\%$                    & $64.94\%$               \\ \hline
Our Method        & $56.44\%$                    & $\mathbf{60.29\%}$                        \\ \hline
\end{tabular}
\vspace{-0.4cm}
\caption{\label{tab:pgd-cifar10}Standard Error and PGD attack success rate on the CIFAR-10 dataset for different state-of-the-art approaches. The chosen $\epsilon$ for the PGD attack equals $8/255$.}
\end{table}

\vspace{-0mm}
\section{Conclusion}
\vspace{-4mm}
We improved the trade-off between standard accuracy and robust 
verifiable accuracy of the shallow and deep neural networks by introducing a training mechanism for networks with multiple abstain classes. We observed that increasing the number of abstain classes results in the ``model degeneracy'' phenomenon where not all abstain classes are utilized, and regular training can lead to solutions with poor performance in terms of standard and robust verified accuracy. 
To avoid the model degeneracy when the number of abstain classes is large, we propose a regularizer scheme penalizing the network if it does not utilize all abstain classes in balance. Our experiments demonstrate the superiority of the trained shallow and deep networks over state-of-the-art approaches on MNIST and CIFAR-10 datasets.
We have used multiple detection classes to improve the performance of the verifiable neural networks. However, the application of multiple detection classes can be beyond such networks for detecting out-of-distribution samples or specific types of adversarial attacks. 

\vspace{-3mm}
\section{Acknowledgment}
\vspace{-4mm}
The work of SB and MR was supported in part  by the AFOSR Young Investigator Program award, by the NSF CAREER award $\#$2144985, a gift from the USC-Meta Center for Research and Education in AI and Learning and by a gift from 3M.

\newpage
\bibliographystyle{abbrvnat}
\bibliography{references}

\newpage
\appendix
\onecolumn
\section{Implementation Details}
\label{appendix: implementation_details}
In table~\ref{tab:structure}, we demonstrate the structure of the deep networks used in experiments of Table~\ref{tab:deep-network}.     
\begin{table}[ht]
\centering
\label{tab:structure}
\begin{tabular}{|c|}
\hline
\textbf{Network Layers} \\ \hline
Conv 64 3×3             \\ \hline
Conv 64 3×3             \\ \hline
Conv 128 3×3            \\ \hline
Conv 128 3×3            \\ \hline
Fully Connected 512     \\ \hline
Linear 10               \\ \hline
\end{tabular}
\vspace{3mm}
\caption{Standard and Robust Verified error of state-of-the-art approaches on CIFAR-10 dataset.}
\end{table}

\begin{enumerate}
    \item For MNIST, we train on a single Nvidia V100 GPU for $100$ epochs with batch sizes of 100. The total number of training steps is 60K. We decay the learning rate by $10\times$ at steps 15K and 25K. We use warm-up and ramp-up duration of 2K and 10K steps, respectively. We do not use any data augmentation techniques and use full $28 \times 28$ images without any normalization.
    
    \item CIFAR-10, we train for $3200$ epochs with batch sizes of $1600$. The total number of training steps is 100K. We decay the learning rate by 10× at steps 60K and 90K. We use warm-up and ramp-up duration of 5K and 50K steps, respectively. During training, we add random translations and flips, and normalize each image channel (using the channel statistics from the train set).
\end{enumerate}

\section{Bregman-Divergence Method for Optimizing a Convex Function Over a Probability Simplex}
\label{appendix: bregman_divergence}

In this section, we use the Bregman divergence method to optimize a convex optimization problem over a probability simplex. Let $\boldsymbol{\eta}$ be a vector of $n$ elements. We aim to minimize the following constrained optimization problem where $J$ is a convex function with respect to $\boldsymbol{\eta}$:

\begin{equation}
\label{eq: general_constrained}
    \min_{\eta_1, \dots, \eta_n} J(\eta_1, \dots, \eta_n) \quad \textrm{subject to } \quad \sum_{i=1}^n \eta_i = 1, \quad \eta_i \geq 0 \quad \forall i = 1, \dots, n. 
\end{equation}

To solve the above problem, we define the Bregman distance function as: 
\begin{equation*}
    B(\bx, \by) = \gamma(\bx)  - \gamma(\by) - \langle \nabla \gamma(\bx), \bx - \by \rangle  
\end{equation*}
where $\gamma$ is a strictly convex function. For this specific problem where the constraint is over a probability simplex, we choose $\gamma (\bx) = \sum_{i=1}^n x_i \log(x_i)$. Thus:

\begin{equation*}
    B(\bx, \by) = \sum_{i=1}^n x_i \log (\frac{x_i}{y_i})
\end{equation*}

One can rewrite problem~\ref{eq: general_constrained} as:

\begin{equation}
\label{eq: general_constrained_rewrite}
    \min_{\eta_1, \dots, \eta_n} J(\eta_1, \dots, \eta_n) + \mathcal{I}_{\mathbb{P}}(\boldsymbol{\eta})
\end{equation}
where $\mathbb{P} = $. Applying the proximal gradient descent method to the above problem, we have:

\begin{align}
    \boldsymbol{\eta}^{r+1} &= \argmin_{\boldsymbol{\eta}} \mathcal{I}_{\mathbb{P}} (\boldsymbol{\eta}) + \langle \nabla J(\boldsymbol{\eta}), \boldsymbol{\eta} - \boldsymbol{\eta}^i \rangle + \frac{1}{2\nu} B(\boldsymbol{\eta}, \boldsymbol{\eta}^i) \\ &= \argmin_{\boldsymbol{\eta}} \sum_{i=1}^n \frac{\partial J(\boldsymbol{\eta}^r)}{\partial \eta_i} (\eta_i - \eta_i^r) + \frac{1}{2\nu} \Big(\sum_{i=1}^n \eta_i \log(\eta_i) -  \sum_{i=1}^n \frac{\partial \gamma(\boldsymbol{\eta}_i^r)}{\partial \eta_i} (\eta_i - \eta_i^r) \Big)
\end{align}

By simplifying the above problem, it turns to:

\begin{align}
    \boldsymbol{\eta}^{r+1} = \argmin_{\boldsymbol{\eta}} & \sum_{i=1}^n \eta_i (\frac{\partial J(\boldsymbol{\eta}^r)}{\partial \eta_i} - \frac{1}{2\nu} \log(\eta_i^r) - \frac{1}{2\nu}) + \frac{1}{2\nu} \sum_{i=1}^n \eta_i \log(\eta_i) \\ &\textrm{subject to } \quad \sum_{i=1}^n \eta_i = 1, \quad \eta_i \geq 0 \quad \forall i = 1, \dots, n. 
\end{align}

Writing the Lagrangian function of the above problem, we have:

\begin{align}
    \boldsymbol{\eta}^{r+1} = \argmin_{\boldsymbol{\eta}} & \sum_{i=1}^n \eta_i (\frac{\partial J(\boldsymbol{\eta}^r)}{\partial \eta_i} - \frac{1}{2\nu} \log(\eta_i^r) - \frac{1}{2\nu}) + \frac{1}{2\nu} \sum_{i=1}^n \eta_i \log(\eta_i) + \lambda^{*} (\sum_{i=1}^n \eta_i - 1) \\ \nonumber & \textrm{subject to } \quad \eta_i \geq 0 \quad \forall i = 1, \dots, n. 
\end{align}

By taking the derivative with respect to $\eta_i$ and using the constraint $\sum_{i=1}^n \eta_i = 1$, it can be shown that:

\begin{equation}
\label{eq: update_rule}
\eta_i^{r+1} = \frac{\eta_i^r \exp(-2\nu \nabla J(\boldsymbol{\eta})_i )}{\sum_{j=1}^n \eta_j^r \exp(-2\nu \nabla J(\boldsymbol{\eta})_j )}    
\end{equation}

We use the update rule~\eqref{eq: update_rule} in Algorithm~\ref{alg: bregman divergence} and Algorithm~\ref{alg: beta_crown_multiple_abstain} to obtain the optimal $\boldsymbol{\eta}$ at each iteration.

\section{Proof of Theorems} 
\label{app: min_max}

In this section, we prove Theorem~\ref{thm: verification_problem_lower_bound} and Theorem~\ref{thm: abstain_loss}.

\textbf{Proof of Theorem}~\textbf{\ref{thm: verification_problem_lower_bound}}:
Starting from Equation~\ref{eq: verification_min_max_multi_abstain}, we can equivalently formulate it as:

\begin{equation}
\label{eq: eq1}
    \min_{\bz \in \mathcal{Z}(\bx_0, \epsilon)} \max (\bc_{yk} ^T \bz,  \bc_{a_1 k} ^T \bz, \dots, \bc_{a_M k} ^T \bz) = \min_{\bz \in \mathcal{Z}(\bx_0, \epsilon)} \max_{\{\eta_0, \dots, \eta_M\} \in \mathcal{P}} \quad  c_k(\boldsymbol{\eta}) ^T \bz.
\end{equation}

Note that the maximum element of the left-hand side can be obtained by setting its corresponding $\eta$ coefficient to $1$ on the right-hand side. Conversely, any optimal solution to the right hand is exactly equal to the maximum element of the left-hand side. According to the min-max equality (duality), when the minimum and the maximum problems are interchanged, the following inequality holds:

\begin{align}
\label{eq: ineq1}
    \min_{\bz \in \mathcal{Z}(\bx_0, \epsilon)} \max_{\{\eta_0, \dots, \eta_M\} \in \mathcal{P}} \quad  &\eta_0 \bc_{yk} ^T \bz + \eta_1 \bc_{a_1 k} ^T \bz + \dots +  \eta_M \bc_{a_M k} ^T \bz \geq \nonumber \\ \max_{\{\eta_0, \dots, \eta_M\} \in \mathcal{P}} \min_{\bz \in \mathcal{Z}(\bx_0, \epsilon)} \quad  &\eta_0 \bc_{yk} ^T \bz + \eta_1 \bc_{a_1 k} ^T \bz + \dots +  \eta_M \bc_{a_M k} ^T \bz.
\end{align}

Moreover, by the definition of upper-bounds and lower-bounds presented in~\cite{gowal2018effectiveness}, $\mathcal{Z}(\bx_0, \epsilon)$ is a subset of $\underline{\bz}_{L} \leq \bz \leq \bar{\bz}_{L}$. Thus:

\begin{align}
\label{eq: ineq2}
     \max_{\{\eta_0, \dots, \eta_M\} \in \mathcal{P}} \min_{\bz \in \mathcal{Z}(\bx_0, \epsilon)} \quad  &\eta_0 \bc_{yk} ^T \bz + \eta_1 \bc_{a_1 k} ^T \bz + \dots +  \eta_M \bc_{a_M k} ^T \bz \geq \nonumber \\ \max_{\{\eta_0, \dots, \eta_M\} \in \mathcal{P}} \min_{\underline{\bz}_{L} \leq \bz \leq \bar{\bz}_{L}} \quad  &\eta_0 \bc_{yk} ^T \bz + \eta_1 \bc_{a_1 k} ^T \bz + \dots +  \eta_M \bc_{a_M k} ^T \bz.
\end{align}

Combining Equality~\eqref{eq: eq1} with~\eqref{eq: ineq1} and~\eqref{eq: ineq2}, we have:
\begin{equation}
\label{eq: res1}
    \min_{\bz \in \mathcal{Z}(\bx_0, \epsilon)} \max (\bc_{yk} ^T \bz,  \bc_{a_1 k} ^T \bz, \dots, \bc_{a_M k} ^T \bz) \geq \max_{\{\eta_0, \dots, \eta_M\} \in \mathcal{P}} \min_{\underline{\bz}_{L} \leq \bz \leq \bar{\bz}_{L}} \quad  c_k(\boldsymbol{\eta}) ^T \bz.
\end{equation}

Since $\bz_L = \bW_L \bz_{L-1} + \bb_L$, the right-hand-side of the above inequality can be rewritten as:
\begin{equation*}
    \min_{\bz \in \mathcal{Z}(\bx_0, \epsilon)} \max (\bc_{yk} ^T \bz,  \bc_{a_1 k} ^T \bz, \dots, \bc_{a_M k} ^T \bz) \geq \max_{\boldsymbol{\eta} \in \mathcal{P}} \min_{\underline{\bz}_{L-1} \leq \bz \leq \bar{\bz}_{L-1}} \quad  c(\boldsymbol{\eta})^T (\bW_L \bz + \bb_L),
\end{equation*}
which is exactly the claim of Theorem~\ref{thm: verification_problem_lower_bound}.

\textbf{Proof of Theorem}~\textbf{\ref{thm: abstain_loss}}:
For the simplicity of the presentation, assume that $a_0 = y$. Partition the set of possible values of $\bz_L$ in the following sets:

\begin{equation*}
    \hat{\mathcal{Z}}_{a_i} = \{ \bz_L | [\bz_L]_{a_i} \geq [\bz_L]_{a_j} \: \forall j \neq i \}
\end{equation*}

If $\bz_L \in \hat{\mathcal{Z}}_{a_i}$, then:

\begin{align*}
    [\bz_L]_{a_i} - [\bz_L]_{k} &\geq [\bz_L]_{a_j} - [\bz_L]_{k} \quad \forall j \neq i \Rightarrow [\bz_L]_{a_i} - [\bz_L]_{k} \\ &= \max_{i=0, \dots, M} \{[\bz_L]_{a_i} - [\bz_L]_{k}\} = \max_{i \in \{ 0, \dots, M\}} \{\bc_{a_i, k}^T \bz_L\} 
\end{align*}

Thus:

\begin{align}
\label{eq: ineq1_thm2}
    [\bz_L]_{a_i} - [\bz_L]_{k} = \max_{i=0, \dots, M} \{\bc_{a_i, k}^T \bz_L\} &\geq \min_{\bz_L \in \mathcal{Z}(\bx_0, \epsilon)} \max_{i=0, \dots, M} \{\bc_{a_i, k}^T \bz_L\} \nonumber \\
    &= \min_{\bz_{L-1} \in \mathcal{Z}_{L-1}(\bx_0, \epsilon)} \max_{i=0, \dots, M} \{\bc_{a_i, k}^T (\bW_L \bz_{L-1}  + \bb_L)\} \nonumber \\  &\geq \min_{\underline{\bz} \leq \bz_{L-1} \leq \bar{\bz}} \max_{i=0, \dots, M} \{\bc_{a_i, k}^T (\bW_L \bz_{L-1}  + \bb_L)\} \nonumber \\  &= \min_{\underline{\bz} \leq \bz_{L-1} \leq \bar{\bz}} \max_{\boldsymbol{\eta} \in \mathbb{P}} c(\boldsymbol{\eta})^T (\bW_L\bz_{L-1}  + \bb_L) 
\end{align}

Note that the second inequality holds since the minimum is taken over a larger set in the right hand side of the inequality. Using the min-max inequality:
\begin{align}
\label{eq: ineq2_thm2}
    \min_{\underline{\bz} \leq \bz_{L-1} \leq \bar{\bz}} \max_{\boldsymbol{\eta} \in \mathbb{P}} c(\boldsymbol{\eta})^T (\bW_L \bz_{L-1} + \bb_L) \geq \max_{\boldsymbol{\eta} \in \mathbb{P}} \min_{\underline{\bz} \leq \bz_{L-1} \leq \bar{\bz}}  c(\boldsymbol{\eta})^T (\bW_L \bz_{L-1}  + \bb_L) = -J_k(\boldsymbol{\eta})
\end{align}

Combining~\eqref{eq: ineq1_thm2} and~\eqref{eq: ineq2_thm2}, and multiplying both sides by $-1$, we obtain:
\begin{align}
\label{eq: thm_part1}
    [\bz_L]_{k} - [\bz_L]_{a_i} \leq J_k(\boldsymbol{\eta})  
\end{align}

On the other hand:
\begin{align}
\label{eq: thm_part2}
&\max_{\| \boldsymbol{\delta} \|_\infty \leq \epsilon} \min_{m=0, \dots, M} \ell_{\textrm{xent} \setminus \mathcal{A}_m} \Big(\bz_L(\bx + \delta),  a_m  \Big) \nonumber \\ &\leq \max_{\| \boldsymbol{\delta} \|_\infty  \leq \epsilon} \ell_{\textrm{xent} \setminus \mathcal{A}_i} \Big(\bz_L(\bx + \delta),  a_i  \Big) \nonumber \\ & \leq \max_{\underline{\bz}_{L-1} \leq \bz \leq \bar{\bz}_{L-1}} \ell_{\textrm{xent}\setminus \mathcal{A}_i}  (\bz_{L}) \quad \textrm{s.t.} \quad \bz_L = \bW_L \bz_{L-1} + \bb_L.
\end{align} 

Moreover, by the property of the cross-entropy loss, we have:
\begin{align}
\label{eq: thm_part3}
\small    
\ell_{\textrm{xent}\setminus \mathcal{A}_i}  (\bz_{L}) = \ell_{\textrm{xent}\setminus \mathcal{A}_i}  (\bz_{L} - [\bz_{L}]_{a_i}\mathbf{1}) 
\end{align}

Combining~\eqref{eq: thm_part1}, \eqref{eq: thm_part2} and~\eqref{eq: thm_part3}, we have:
\begin{align}
    & \max_{\| \boldsymbol{\delta} \|_\infty \leq \epsilon} \min_{m=0, \dots, M} \ell_{\textrm{xent} \setminus \mathcal{A}_m} \Big(\bz_L(\bx + \delta),  a_m  \Big) \nonumber \\ &\leq   \max_{\underline{\bz}_{L-1} \leq \bz_{L-1} \leq \bar{\bz}_{L-1}} \ell_{\textrm{xent}\setminus \mathcal{A}_i}  (\bz_{L}) \quad \textrm{s.t.} \quad \bz_L = \bW_L \bz_{L-1}  + \bb_L. \nonumber \\ &= \max_{\underline{\bz}_{L-1} \leq \bz_{L-1} \leq \bar{\bz}_{L-1}} \ell_{\textrm{xent}\setminus \mathcal{A}_i}  (\bz_{L} - [\bz_{L}]_{a_i}\mathbf{1}) \quad \textrm{s.t.} \quad \bz_L = \bW_L \bz_{L-1} \nonumber \\ &\leq  \max_{\underline{\bz}_{L-1} \leq \bz_{L-1} \leq \bar{\bz}_{L-1}} \ell_{\textrm{xent}\setminus \mathcal{A}_i}  (J_k(\boldsymbol{\eta}), a_i) \nonumber \\ &= \max_{\underline{\bz}_{L-1} \leq \bz_{L-1} \leq \bar{\bz}_{L-1}} \ell_{\textrm{xent}\setminus \mathcal{A}_0}  (J_k(\boldsymbol{\eta}), a_0) \nonumber 
\end{align}

Summing up over all data points, the desired result is proven.

\section{Details of $\beta$-Crown}
\label{appendix: beta_crown}
In this section, we show how $\beta$-crown sub-problems can be obtained for neural networks without abstain classes and with multiple abstain classes respectively. Before proceeding, let us have a few definitions and lemmas.

\begin{lemma}
\label{app: lemma1}
\textbf{\citep[Theorem~15]{zhang2019towards}} Given two vectors $\mathbf{u}$ and $\mathbf{v}$, the following inequality holds:

\begin{gather*}
    \bv^\top \textup{ReLU}(\bu) \geq \bv^\top \bD_{\boldsymbol{\alpha}} \bu + \bb^\prime, 
\end{gather*}

where $\bb^\prime$ is a constant vector and $\bD_{\boldsymbol{\alpha}}$ is a diagonal matrix containing $\boldsymbol{\alpha}_j$'s as free parameters: 

\begin{equation}
\bD_{j,j}(\boldsymbol{\alpha}) = \begin{cases}
1, & \text{if $\underline{\bz}_j \geq 0$} \\
0, & \text{if $\bar{\bz}_j \leq 0$} \\
\boldsymbol{\alpha}_j, & \text{if $\bar{\bz}_j > 0 > \underline{\bz}_j$ and $\bv_{j} \geq 0$} \\
\frac{\bar{\bz}_j}{\bar{\bz}_j - \underline{\bz}_j}, & \text{if $\bar{\bz}_j > 0 > \underline{\bz}_j$ and $\bv_j < 0$}, 
\end{cases}
\label{eq:relax_single_d}
\end{equation}

\end{lemma}

\begin{definition}
The recursive function $\Omega(i, j)$ is defined as follows~\citep{wang2021beta}:

\begin{equation*}
    \Omega(i, i) = \boldsymbol{I}, \quad 
    \Omega(i, j) = \boldsymbol{W}_{i} \boldsymbol{D}_{i-1}(\boldsymbol{\alpha}_{i-1}) \Omega(i-1, j) 
\end{equation*}

\end{definition}

$\beta$-crown defines a matrix $\bS$ for handling splits through the branch-and-bound process. The multiplier(s) $\boldsymbol{\beta}$ determines the branching rule.

\begin{equation}
\bS_{i}[j][j] = \begin{cases}
-1, & \text{if split $\bz_i[j] \geq 0$} \\
1, & \text{if split $\bz_i[j] < 0$} \\
0, & \text{if no split $\bar{\bz}_j$}, 
\end{cases}
\label{eq:relax_single_S}
\end{equation}

Thus, the verification problem of $\beta$-crown is formulated as:

\begin{equation}
    \min_{\bz in \mathcal{Z}} \bc^T \big(\bW^L \textrm{ReLU}(\bz_{L-1}) + \bb_{L-1} \big) \geq 
    \min_{\bz in \mathcal{Z}} \quad \max_{\boldsymbol{\beta}_{L-1}} \: \bc^T \big(\bW^L \bD_{L-1} \bz_{L-1} + \bb_{L-1} \big) + \boldsymbol{\beta}_{L-1}^\top \bS_{L-1} 
\end{equation}

Having these definitions, we can write $\bP, \bq, \ba,$ and $\bd$ explicitly as functions of $\boldsymbol{\alpha}$ and $\boldsymbol{\beta}$. $\bP \in \mathbb{R}^{d_0 \times (\sum_{i=1}^{L-1} {d_i})}$ is a block matrix $\bP:=\left [ \bP_{1}^\top \ \bP_{2}^\top \  \cdots \ \bP_{L-1}^\top \right ]$, $\bq \in \mathbb{R}^{\sum_{i=1}^{L-1} d_i}$ is a vector $\bq := \left [ \bq_{1}^\top \ \cdots \ \bq_{L-1}^\top \right ]^\top$. Moreover:

\begin{equation*}
\ba = \left [ \Omega(L, 1) \bW_{1} \right ]^\top \in \mathbb{R}^{d_0 \times 1},
\end{equation*}

\begin{equation*}
\bP_{i} = \bS_{i} \Omega(i,1) \bW_{1} \in \mathbb{R}^{d_i \times d_0}, \quad \forall \ 1 \leq i \leq L-1
\end{equation*}

\begin{equation*}
\bq_{i} = \sum_{k=1}^{i} \bS_{i} \Omega(i, k) \bb_{k} + \sum_{k=2}^{i} \bS_{i} \Omega(i, k) \bW_{k} \underline{\bb}_{k-1} \in \mathbb{R}^{d_i}, \quad \forall \ 1 \leq i \leq L-1
\end{equation*}

\begin{equation*}
\bd = \sum_{i=1}^{L} \Omega(L, i) \bb_{i} + \sum_{i=2}^{L} \Omega(L, i) \bW_{i} \underline{b}_{i-1} 
\end{equation*}

\begin{equation*}
\underline{b}_{i} =  \begin{cases}
1, & \text{if $\underline{\bz}_j \geq 0$} \\
0, & \text{if $\bar{\bz}_j \leq 0$} \\
\boldsymbol{\alpha}_j, & \text{if $\bar{\bz}_j > 0 > \underline{\bz}_j$ and $\bv_{j} \geq 0$} \\
\frac{\bar{\bz}_j}{\bar{\bz}_j - \underline{\bz}_j}, & \text{if $\bar{\bz}_j > 0 > \underline{\bz}_j$ and $\bv_j < 0$}, 
\end{cases}
\end{equation*}

Now we extend the definition of $g$ for the network consisting of multiple abstain classes. Let $\bar{\bz}$ be the pre-activation value of vector $z$ before applying the ReLU function. We aim to solve the following verification problem:

\begin{align} 
 \min_{\bz_{L-1} \in \mathcal{Z}_{L-1}(\bx_0, \epsilon)} \quad \max_{\boldsymbol{\eta} \in \mathcal{P}} \nonumber  c_k(\boldsymbol{\eta})^T (\bW_L \bz_{L-1} + \bb_L).
\label{eq: verification_min_max_original}
\end{align}

Applying Lemma~\ref{app: lemma1} to the above problem, we have:

\begin{align} 
 \min_{\bz_{L-1} \in \mathcal{Z}_{L-1}(\bx_0, \epsilon)} \quad & \max_{\boldsymbol{\eta} \in \mathcal{P}} \nonumber  c_k(\boldsymbol{\eta})^T \Big(\bW_L \bz_{L-1} + \bb_L \Big) \\ \leq \min_{\bz_{L-1} \in \mathcal{Z}_{L-1}(\bx_0, \epsilon)} \quad & \max_{\boldsymbol{\eta} \in \mathcal{P}} \nonumber  c_k(\boldsymbol{\eta})^T \Big(\bW_L \bD_{L-1} \big( \boldsymbol{\alpha}_{L-1} \big) \hat{\bz}_{L-1} + \bb_L \Big)
\end{align}

Adding the $\beta$-crown Lagrangian multiplier to the above problem, it turns to:
\begin{align} 
\small
  & \min_{\bz_{L-1} \in \mathcal{Z}_{L-1}(\bx_0, \epsilon)} \quad \max_{\boldsymbol{\eta} \in \mathcal{P}} \quad c_k(\boldsymbol{\eta})^T \Big(\bW_L \bD_{L-1} \big( \boldsymbol{\alpha}_{L-1} \big) \hat{\bz}_{L-1} + \bb_L \Big) \leq \nonumber \\ 
 & \min_{\bz_{L-1} \in \mathcal{Z}_{L-1}(\bx_0, \epsilon)} \quad \max_{\boldsymbol{\eta} \in \mathcal{P}, \boldsymbol{\alpha}_{L-1}, \boldsymbol{\beta}_{L-1}} \quad c_k(\boldsymbol{\eta})^T \Big( \bW_L \bD_{L-1} ( \boldsymbol{\alpha}_{L-1}) \bz_{L-1} + \bb_L \Big) + \boldsymbol{\beta}_{L-1}^\top \bS_{L-1}  \bz_{L-1} \nonumber \\ & 
 \leq \max_{\boldsymbol{\alpha}_{L-1}, \boldsymbol{\beta}_{L-1}} \min_{\bz_{L-1} \in \mathcal{Z}_{L-1}(\bx_0, \epsilon)} \quad \max_{\boldsymbol{\eta} \in \mathcal{P}} \Big( c_k(\boldsymbol{\eta})^T \bW_L \bD_{L-1}( \boldsymbol{\alpha}_{L-1}) + \boldsymbol{\beta}_{L-1}^\top \bS_{L-1} \Big) \hat{\bz}_{L-1} \nonumber \\ & + c_k(\boldsymbol{\eta})^T \bb_L 
 =  \max_{\boldsymbol{\alpha}_{L-1}, \boldsymbol{\beta}_{L-1}}  \min_{\bz_{L-1} \in \mathcal{Z}_{L-1}(\bx_0, \epsilon)} \quad \max_{\boldsymbol{\eta} \in \mathcal{P}} \Big( c_k(\boldsymbol{\eta})^T \bW_L \bD_{L-1}( \boldsymbol{\alpha}_{L-1}) \nonumber \\ &+ \boldsymbol{\beta}_{L-1}^\top \bS_{L-1} \Big) \Big(\bW_{L-1} \bz_{L-2} + \bb_{L-1} \Big) + c_k(\boldsymbol{\eta})^T \bb_L \nonumber
\end{align}

Replace the definition of $\bA^{(i)}$ in \citep[Theorem~3.1]{wang2021beta} with the following matrix and repeat the proof. 

\begin{equation}
\bA^{(i)} = \begin{cases}
c_k(\boldsymbol{\eta})^T \bW_L, & \text{if} \quad i = L-1 \\
\Big( \bA^{(i+1)}\bD_{i+1} (\boldsymbol{\alpha}_{i+1}) + \boldsymbol{\beta}_{i+1}^\top \bS_{i+1} \Big)\bW_{i+1}, & \text{if} \quad 0 \leq i \leq L-2 
\end{cases}
\end{equation}

Note that the definition of $\bd$ will be changed in the following way:

\begin{equation*}
\bd = c_k(\boldsymbol{\eta})^T \bb_L + \sum_{i=1}^{L} \Omega(L, i) \bb_{i} + \sum_{i=2}^{L} \Omega(L, i) \bW_{i} \underline{b}_{i-1} 
\end{equation*}

Moreover, $\Omega(L, j) = c_k(\boldsymbol{\eta})^T \boldsymbol{W}_{L} \boldsymbol{D}_{L-1}(\boldsymbol{\alpha}_{L-1}) \Omega(L-1, j)$. The rest of the definitions remain the same.

\section{Derivation of equation~\eqref{eq: beta_crown_subproblem_multiple_abstain}}
\label{appendix: derivation_G_inequality}

In this section, we show how to derive Equation~\ref{appendix: derivation_G_inequality}. 

\begin{align}
    & \min_{\bz_L \in \mathcal{Z}(\bx, \epsilon)} \quad \max \{ \bc_{yk}^T \bz_L, \bc_{a_1 k}^T \bz_L, \dots, \bc_{a_M k}^T \bz_L\} \nonumber \\ &= \min_{\bz_L \in \mathcal{Z}(\bx, \epsilon)} \quad \max_{\boldsymbol{\eta} \in \mathcal{P}} \sum_{i=0}^M \eta_i c_{a_i k}^T \bz_L \nonumber \\ &\geq \max_{\boldsymbol{\eta} \in \mathcal{P}} \quad \min_{\bz_L \in \mathcal{Z}(\bx, \epsilon)} \sum_{i=0}^M \eta_i c_{a_i k}^T \bz_L \nonumber \\ 
    &\geq \max_{\boldsymbol{\eta} \in \mathcal{P}} \quad \max_{\boldsymbol{\alpha}, \boldsymbol{\beta} \geq 0} \eta_i c_{a_i k}^T \bz_L \nonumber \\ &= \max_{\boldsymbol{\alpha}, \boldsymbol{\beta} \geq 0, \boldsymbol{\eta} \in \mathcal{P}}
    \Big( \sum_{i=0}^M \eta_i g_i(\bx_0, \boldsymbol{\alpha}, \boldsymbol{\beta})  \triangleq G(\bx_0, \boldsymbol{\alpha}, \boldsymbol{\beta}, \boldsymbol{\eta}) \Big) \nonumber
\end{align}
\section{A simple example on the benefits and pitfalls of having multiple abstain classes}
\label{appendix: multi_abstain_motivation}

In this example, we provide a simple toy example illustrating:
\begin{enumerate}
    \item How adding multiple abstain classes can improve the detection of adversarial examples.
    
    \item How detection with multiple abstain classes may suffer from a ``model degeneracy" phenomenon.
\end{enumerate}

\begin{figure*}
\begin{center}
\centerline{\includegraphics[width=0.6\columnwidth]{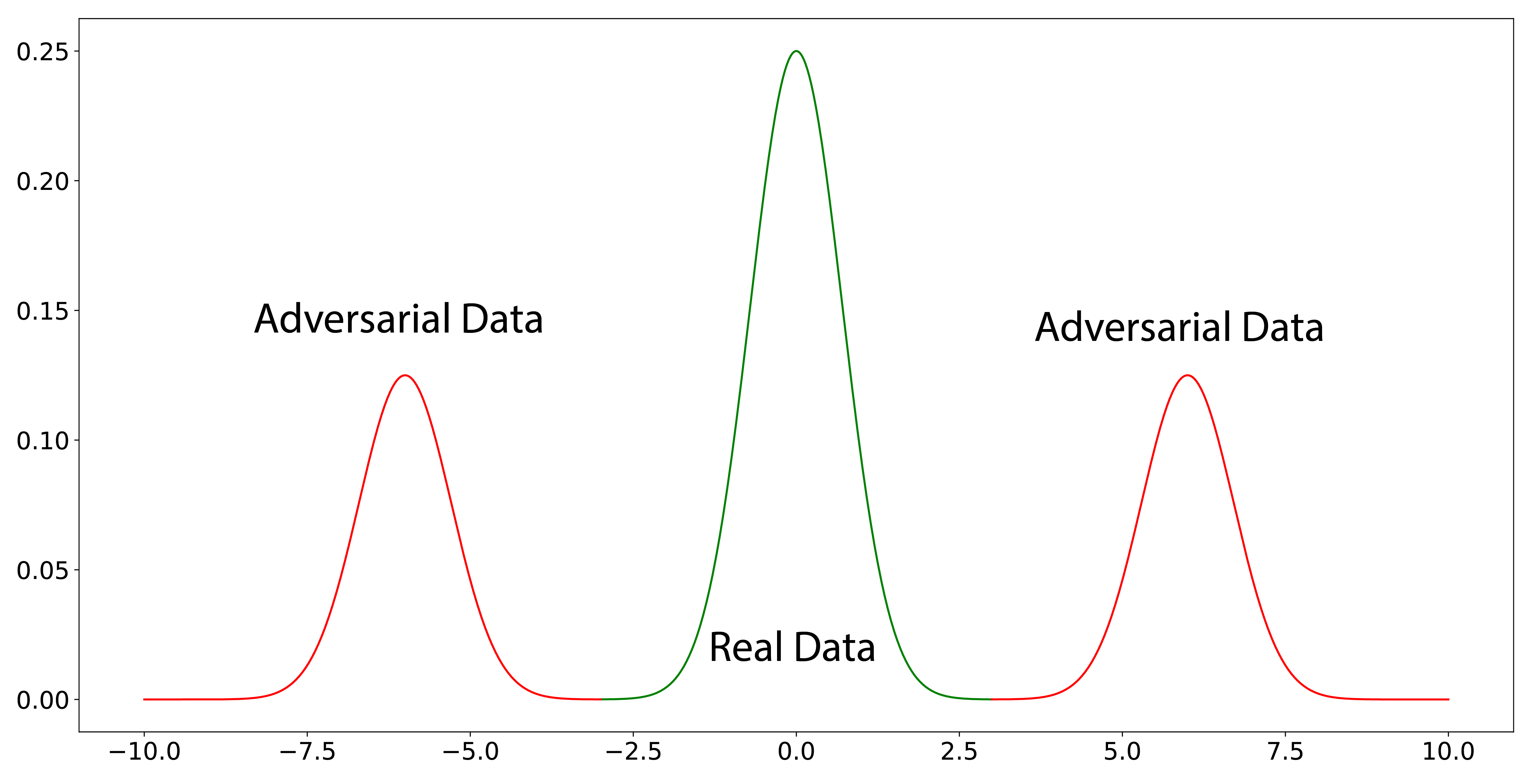}}
\caption{Distribution of adversarial and real data described in the example. While one linear classifier cannot separate the adversarial (red section) and real (green section) data points, two detection classes can detect adversarial examples.}
\label{fig: supp}
\end{center}
\end{figure*}

\textbf{Example:} Consider a simple one-dimensional data distributed where the read data is coming from the Laplacian distribution with probability density function $P_r(X=x) = \frac{1}{2} \exp(-|x|)$. Assume that the adversary samples are distributed according to the probability density function $P_a(X=x) = \frac{1}{4} (\exp(-|x-10|) + \exp(-|x+10|)$. Assume that $\frac{1}{3}$ data is real, and $\frac{2}{3}$ is coming from adversary. The adversary and the real data are illustrated in Fig~\ref{fig: supp}.

Consider a binary neural network classifier with no hidden layer for detecting adversaries. Specifically, the neural network has two weight vectors $w^r$ and $w^a$, and the bias values $b^r$ and $b^a$. The network classifies a sample $\bx$ as "real" if $w^r x + b^r > w^a x + b^a$; otherwise, it classifies the sample as out-of-distribution/abstain. The misclassification rate of this classifier is given by:
\begin{align*}
    P(\textrm{error}) &= \frac{1}{3} P_{x \sim P_r}(w^a x + b^a > w^r x + b^r) + \frac{2}{3} P_{x \sim P_a}(w^a x + b^a < w^r x + b^r) \\
    &= \frac{1}{3} P_{x \sim P_r}( x > \frac{b^r - b^a}{w^a - w^r}) + \frac{2}{3} P_{x \sim P_a}( x < \frac{b^r - b^a}{w^a - w^r}), 
\end{align*}
where due to symmetry and scaling invariant, without loss of generality, we assumed that $w^a - w^r > 0$. Let $t = \frac{b^r - b^a}{w^a - w^r}$. Therefore,
\begin{equation}
    P(\textrm{error}) = \frac{1}{3} \int_{t}^{+\infty} \frac{1}{2} \exp(-|x|) dx + \frac{2}{3} \int_{-\infty}^{t} \frac{1}{4} (\exp(-|x-10|) + \exp(-|x+10|) dx 
\end{equation}
Thus, to find the optimal classifier, we require to determine the optimal $t$ minimizing the above equation. One can numerically verify that the optimal $t$ is given by $t^* = 5$, leading to the minimum misclassification rate of $\approx  0.34$. This value is the optimal misclassification rate that can be achieved by our single abstain class neural network.

Now consider a neural network with two abstain classes. Assume that the weights and biases corresponding to the abstain classes are $w_1^a, w_2^a, b_1^a, b_2^a,$ and the weight and bias for the real class is given by $w_r$ and $b_r$. A sample $x$ is classified as a real example if and only if both of the following conditions hold:
\begin{align}
\label{conditions: two_abstain}
    w^r x + b_r > w_1^a x + b_1^a \\
    w^r x + b_r > w_2^a x + b_2^a,
\end{align}
otherwise, it is classified as an adversarial (out of distribution) sample. The misclassification rate of the such classifier is given by:
\begin{equation}
\label{eq: misclassification_rate}
P(\textrm{error}) = \frac{1}{3} P_{x \sim P_c} (\textrm{Conditions}~\eqref{conditions: two_abstain} \: \textrm{hold}) + \frac{2}{3} P_{x \sim P_a} (\textrm{Conditions}~\eqref{conditions: two_abstain} \: \textrm{do not hold}) 
\end{equation}

\textbf{Claim 1:} The point $w_1^a = -1, w_2^a = 1, b_1^a = b_2^a = 0, b^r = 5, w^r = 0$ is a global minimum of~\eqref{eq: misclassification_rate} with the optimum misclassification rate less than $0.1$.

\textbf{Proof:} Define $t_1 = - \frac{b_1^a - b_r}{w_1^a - w^r}, t_2 = - \frac{b_2^a - b_r}{w_2^a - w^r}$. Considering all possible sign cases, it is not hard to see that at the optimal point, $w_1^a - w^r$ and $w_2^a - w^r$ have different signs. Without loss of generality, assume that $w_1^a - w^r < 0$ and $w_2^a - w^r > 0$. Then:
\begin{equation}
\label{eq: misclassification_rate_t}
P(\textrm{error}) = \frac{1}{3} P_{x \sim P_c} (x \leq t_1 \lor x \geq t_2) + \frac{2}{3} P_{x \sim P_a} (x \geq t_1 \land x \leq t_2) 
\end{equation}
It is not hard to see that the optimal solution is given by $t_1^{*} = -5$, $t_2^{*} = 5$. Plugging these values in the above equation, we can check that the optimal loss is less than $0.1$. $\blacksquare$

Claim~1 shows that by adding an abstain class, the misclassification rate of the classifier goes down from $0.34$ to below $0.1$. This simple example illustrates the benefit of having multiple abstain classes. Next, we show that by having multiple abstain classes, we are prone to the ``model degeneracy" phenomenon. 

\textbf{Claim 2:} Let $\bar{w}_1^a = \bar{w}_2^a = 1, \bar{b}_1^a = \bar{b}_2^a = 0, \bar{w}^r = 0, \bar{b}^r = 5$. Then, there exists a point $(\tilde{w}, \tilde{b}) = (\tilde{w}_1^a, \tilde{w}_2^a, \tilde{b}_1^a, \tilde{b}_2^a, \tilde{w}^r, \tilde{b}^r)$ such that  $(\tilde{w}, \tilde{b})$ is a local minimum of the loss function in~\eqref{eq: misclassification_rate} and $\| (\tilde{w}, \tilde{b}) - (\bar{w}, \bar{b})\|_2 \leq 0.1$. 

\textbf{Proof:} 
Let $t_1 = - \frac{b_1^a - b_r}{w_1^a - w^r}, t_2 = - \frac{b_2^a - b_r}{w_2^a - w^r}$. Notice that in a neighborhood of point $(\bar{w}, \bar{b})$, we have $w_1^a - w^r > 0$ and $w_2^a - w^r > 0$. Thus, after the loss function in~\eqref{eq: misclassification_rate} can be written as:
 \begin{align*}
    \ell(t_1, t_2) &= \frac{1}{3} P_{x \sim P_c} (x \leq t_1 \lor x \geq t_2) + \frac{2}{3} P_{x \sim P_a} (x \geq t_1 \land x \leq t_2) \\
    &= \frac{1}{3} P_{x \sim P_r} (x \geq \min(t_1, t_2)) + \frac{2}{3} P_{x \sim P_r} (x \leq \min(t_1, t_2)) \\
    &= \frac{1}{3} P_{x \sim P_r} (x \geq z) + \frac{2}{3} P_{x \sim P_r} (x \leq z),
\end{align*}
where $z = \min_{t_1, t_2}$. It suffices to show that the above function has a local minimum close to the point $\bar{z} = 5$ (see~\citep{nouiehed2021learning}). Simplifying $\ell(t_1, t_2)$ as a function of $z$, we have:
\begin{equation*}
    \ell(t_1, t_2) = h(z) = \frac{1}{6} \exp(-z) + \frac{1}{3} - \frac{1}{6} \exp(-z-10) + \frac{1}{6} \exp(z-10) 
\end{equation*}
Plotting $h(z)$ shows that it has a local minimum close to $\bar{z} = 5$. $\blacksquare$

This claim shows that by optimizing the loss, we may converge to the local optimum $(\tilde{w}, \tilde{b})$ where both abstain classes become essentially the same and we do not utilize the two abstain classes fully.

\section{Structure of Neural Networks in Section~\ref{sec:verification}}
\label{appendix: network_structure}
In Section~\ref{sec:verification} we introduced a toy example in the Motivation subsection to show how loser IBP bounds can become when we go from a $2$-layer network to an equivalent $3$-layer network. The structure of the $2$-layer neural networks is as follows:
\begin{equation*}
    \bz_2 (\bx) = \bW_2 \textrm{ReLU}(\bW_1 \bx),  
\end{equation*}
where $\bx$ is the 2-dimensional input, $\bW_1 = \begin{pmatrix}
0.557 & -0.296 & -0.449\\
-0.474 & -0.504 & 0.894\\
-0.0208 & 0.0679 & 0.901
\end{pmatrix}$, and $\bW_2 = \begin{pmatrix}
0.817 & -0.376 & 0.36\\
0.524 & 0.530 & 0.0557\\
0.0753 & 0.191 & 0.744\\
-0.547 & 0.660 & -0.718 
\end{pmatrix}$. 

Note that the input data is $2$ dimensional, but we add an extra one for incorporating bias into $\bW_1$ and $\bW_2$. The chosen $\epsilon$ for each data point equals $1$.

\section{Experiments on Deep Neural Networks}
To compare the performance of our proposed approach to other state-of-the-art methods on deep neural networks, we run the methods on the network with the structure described in Appendix~\ref{appendix: implementation_details}. The results are reported in Table~\ref{tab:deep-network}.

\begin{table}[H]
\centering
\begin{adjustbox}{width=\textwidth}
\begin{tabular}{|c|c|c|c|} 
\hline
$\epsilon$  & Method                                            & Standard Error (\%) & Robust Verified Error (\%)  \\ 
\hline
         & Interval Bound Propagation \citep{gowal2018effectiveness}                                              & 50.51              & 68.44                       \\ 
\cline{2-4}
         & IBP-CROWN \citep{zhang2019towards}                                         & 54.02              & 66.94                       \\ 
\cline{2-4}
$\epsilon_{\textrm{train}} = 8.8/255$  & \citep{balunovic2019adversarial}                & 48.3               & 72.5                        \\ 
\cline{2-4}
         & Single Abstain \citep{sheikholeslamiprovably}                          & 55.60              & 63.63                       \\ 
\cline{2-4}
$\epsilon_{\textrm{test}} = 8/255$    & \textbf{Multiple Abstain Classes (Current Work)}                         & 56.72              & 61.45                       \\ 
\cline{2-4}
         & \textbf{Multiple Abstain Classes (Verified by Beta-crown)} & 56.72              & 57.55                       \\ 
\hline
         & Interval Bound Propagation \citep{gowal2018effectiveness}                                               & 68.97              & 78.12                       \\ 
\cline{2-4}
$\epsilon_{\textrm{train}} = 17.8/255$ & IBP-CROWN \citep{zhang2019towards}                                          & 66.06              & 76.80                       \\ 
\cline{2-4}
         & Single Abstain \citep{sheikholeslamiprovably}                          & 66.37              & 67.92                       \\ 
\cline{2-4}
$\epsilon_{\textrm{test}} = 16/255$   & \textbf{Multiple Abstain Classes (verified by IBP)}                          & 66.25              & 64.57                       \\ 
\cline{2-4}
         & \textbf{Multiple Abstain Classes (Verified by Beta-crown)} & 66.25              & 62.81                       \\
\hline
\end{tabular}
\end{adjustbox}
\caption{\label{tab:deep-network}Standard and Robust Verified error of state-of-the-art approaches on CIFAR-10 dataset.}
\end{table}

\newpage
\vspace{-3mm}
\section{Limitations}
\vspace{-3mm}
The proposed framework for training and verifying joint detector and classifier networks defines the uncertainty set on each sample as an $L_{\infty}$ norm ball. The results can be extended to other $L_p$ norm balls ($L_1$ or $L_2$) by changing the Interval Bound Propagation~\citep{gowal2018effectiveness} procedure to other constraint sets defined by $L_p$ balls. However, the experiments in the paper are performed on the $L_{\infty}$ constraint sets. Furthermore, the networks in the numerical section are trained on MNIST and CIFAR-10 datasets due to the expensive and complex training procedure of \textbf{verifiable} neural networks. The training procedure of the verifiably robust neural networks is yet limited to these datasets, even for the fastest methods such as IBP. Besides, on large-scale datasets with millions of samples and many different classes, the optimal $M$ might be much larger compared to its optimal value of $M=3$ and $M=4$ on MNIST and CIFAR-10 datasets, respectively. Therefore, it is crucial to devise techniques for training large verifiable neural networks in general and networks with multiple detection classes in particular.

\section{Societal Impacts} \label{App:SocietalImpacts}

Given the susceptibility of presently trained neural networks to adversarial examples and out-of-distribution samples, the deployment of such models in critical applications like self-driving cars has been the subject of debate. Ensuring the reliability and safety of neural networks in unpredictable and adversarial environments requires the development of mechanisms that guarantee the models' robustness. Our current study proposes a systematic approach for training and validating neural networks against adversarial attacks. From a broader perspective, establishing verifiable assurances for the performance of artificial intelligence (AI) models alleviates ethical and safety concerns associated with AI systems.

\end{document}